\documentclass[11pt]{article}

\usepackage[final]{acl}
\usepackage{enumitem}
\usepackage{times}
\usepackage{latexsym}
\usepackage{multirow}
\usepackage{tabularx}
\usepackage{booktabs}
\usepackage{xcolor}

\usepackage{amsmath} % 必须引入以支持 gathered 等公式环境
\usepackage{tcolorbox}
\tcbuselibrary{skins, breakable}

\newtcolorbox{promptbox}{
    enhanced, breakable,
    colback=gray!5!white, 
    colframe=black, boxrule=0.75pt, arc=0pt, 
    left=10pt, right=10pt, top=12pt, bottom=10pt,
    title={User Prompt},
    fonttitle=\bfseries\normalsize,
    coltitle=black,
    attach boxed title to top left={xshift=15pt, yshift=-\tcboxedtitleheight/2},
    boxed title style={colback=gray!5!white, frame hidden, left=4pt, right=4pt, bottom=0pt, top=0pt}
}

\newtcolorbox{modelbox}[1]{
    enhanced, breakable,
    colback=white, 
    colframe=black, boxrule=0.75pt, arc=0pt, 
    left=10pt, right=10pt, top=12pt, bottom=10pt,
    title={Model Output: #1},
    fonttitle=\bfseries\normalsize,
    coltitle=black,
    attach boxed title to top left={xshift=15pt, yshift=-\tcboxedtitleheight/2},
    boxed title style={colback=white, frame hidden, left=4pt, right=4pt, bottom=0pt, top=0pt}
}

\usepackage{siunitx}            % 按小数点对齐 + 自动千分位
\definecolor{rowgray}{gray}{0.93}

\definecolor{hlblue}{RGB}{230, 241, 252}   % Acc. 最优：淡蓝
\definecolor{hltok}{RGB}{252, 240, 226}     % Tok. 最少：淡橙
\newcommand{\ha}[1]{\cellcolor{hlblue}\textbf{#1}}  % 最优准确率
\newcommand{\hc}[1]{\cellcolor{hltok}\textbf{#1}}   % 最少 token

\usepackage[table]{xcolor}  % 必需，提供 \cellcolor
\usepackage{booktabs}
\usepackage{bm}   
\usepackage{multirow}
\usepackage{graphicx}
\newcommand{\method}[1]{#1} % 换用无衬线字体

\usepackage[T1]{fontenc}
\usepackage{amssymb}
\usepackage{amsfonts}
\usepackage{amsmath}
\usepackage{booktabs}
\usepackage[utf8]{inputenc}

\usepackage{microtype}

\usepackage{inconsolata}

\usepackage{graphicx}

\title{MI-Distillation: Selecting from Model-Interpolated Instruct-Reasoning Data Spectrum for Chain-of-Thought Distillation}

\author{Yangsong Lan, Renkai Hu, HongKai Zheng, Bo Zhang, \\
  {\bfseries Renzhi Wang, Hongliang Dai, Piji Li$^{\ast}$} \\
  \textsuperscript{\rm 1}College of Artificial Intelligence, \\
  Nanjing University of Aeronautics and Astronautics, Nanjing, China \\
  \textsuperscript{\rm 2}The Key Laboratory of Brain-Machine Intelligence Technology, Ministry of Education, Nanjing, China \\
  \texttt{\{lys2962331781, hongldai, pjli\}@nuaa.edu.cn} \\}

\begin{document}
\maketitle
\renewcommand{\thefootnote}{\fnsymbol{footnote}}
\footnotetext[1]{Corresponding author.}
\renewcommand{\thefootnote}{\arabic{footnote}}
\begin{abstract}
Recent advances in large reasoning models (LRMs) have shown strong performance on complex problems through long chain-of-thought (Long CoT) reasoning. However, distilling such trajectories into smaller student models remains challenging: direct Long CoT supervision often provides limited gains and can be less effective than concise Short CoT rationales. In this work, we investigate this phenomenon from a gradient-centric perspective. Our analysis shows that Long CoT induces larger gradient magnitudes and more concentrated update directions than Short CoT, with this effect becoming more pronounced as student model capacity increases. These findings suggest that effective Long CoT distillation requires balancing the reasoning information density of reasoning trajectories with their distributional alignment to the student model. Motivated by this insight, we propose \textbf{M}odel \textbf{I}nterpolation \textbf{Distillation} (\textbf{MI-Distillation}), a framework that constructs a continuous Instruct-Reasoning data spectrum through model interpolation. To select suitable trajectories from this spectrum, we further introduce \textbf{Seq}uential \textbf{L}earnable \textbf{S}urprisal \textbf{S}core (\textbf{SeqLSS}), which favors reasoning paths that are both informative and learnable for the student. Extensive experiments on reasoning benchmarks show that MI-Distillation consistently improves small model CoT distillation over strong Long CoT baselines.\footnote{Our implementation is available at \href{https://github.com/yslanprime/MI-Distillation}{GitHub}.}
\end{abstract}

\section{Introduction}
Recent advances in large reasoning models (LRMs), including DeepSeek-R1~\cite{guo2025deepseek}, Seed1.5-Thinking~\cite{seed2025seed1}, and Kimi K2.5~\cite{team2026kimi}, have demonstrated strong capabilities across complex problem-solving and code-generation tasks~\cite{he2024olympiadbench,aime25,jain2025livecodebench}. Trained with reinforcement learning from verifiable rewards (RLVR)~\cite{shao2024deepseekmath}, these models exhibit strong test-time reasoning capabilities, and their Long CoT trajectories have further emerged as a valuable source of supervision for supervised fine-tuning (SFT) and cold-start reinforcement learning~\cite{muennighoff2025s1,guo2025deepseek}.

\begin{figure}[t] % [t] 表示优先置顶，[b] 表示置底
    \centering
    \includegraphics[width=\linewidth]{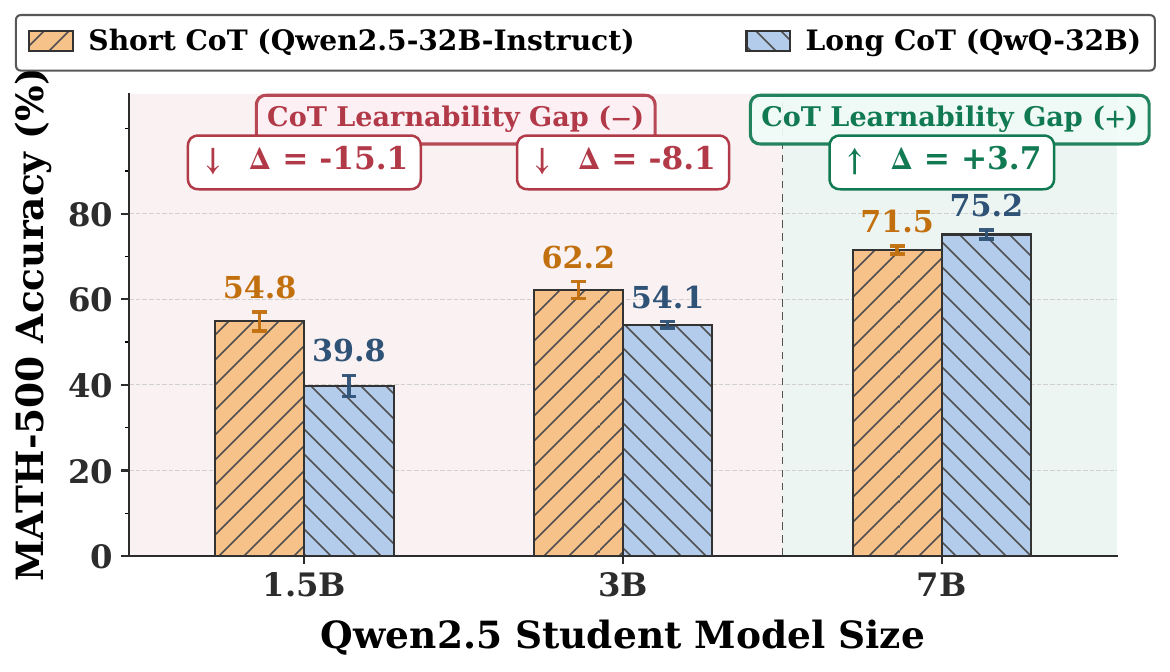}
    \caption{\textbf{Scaling behavior of Short vs. Long CoT distillation on the MATH-500 benchmark.} Accuracy comparison of Qwen2.5 student models fine-tuned under Short and Long CoT supervision. Results indicate a scale-dependent effect: Short CoT provides superior supervision for smaller students (1.5B and 3B), whereas Long CoT marginally overtakes it at the 7B scale.}
    \label{fig:efficiency_comparsion}
    \vspace{-1em}
\end{figure}

However, the high computational cost of LRMs limits their deployment in resource-constrained scenarios~\cite{lan2026crisp,feng2025efficient}, making it essential to transfer their reasoning abilities to smaller models. A natural approach is to fine-tune small models directly on Long CoT traces generated by LRMs, with the expectation that richer reasoning supervision will lead to stronger reasoning performance~\cite{yang2026reasoning,just2025distilling}. Recent evidence shows that small models often fail to benefit from direct Long CoT supervision and may instead achieve better performance when trained on Short CoT data~\cite{li2025small,kim2025their}, as shown in Figure~\ref{fig:efficiency_comparsion}. This highlights the importance of matching reasoning data to the student model in effective distillation.

Motivated by these observations, we investigate the gradient signatures induced by Long and Short CoT across student models of varying scales. Our analysis shows that, compared with Short CoT, Long CoT induces gradients with larger magnitudes and more concentrated directions, and this effect becomes more pronounced as model scale increases. These findings suggest that Long CoT supervision is effective when the student has sufficient capacity to translate verbose reasoning trajectories into coherent and low-dimensional parameter updates. This highlights a key challenge in reasoning distillation: the supervision signal must be not only informative, but also compatible with the student's intrinsic distribution.

Prior work has sought to narrow this gap by mixing Long and Short CoT data~\cite{li2025small} or by adopting curriculum distillation strategies~\cite{jiang2025teach} that gradually reduce the learning difficulty. However, such strategies provide only coarse control over the supervision signal, since the mixing ratios or curriculum schedules are usually manually designed and the resulting trajectories remain limited in diversity. This raises two central questions: (1) How can we construct more fine-grained reasoning trajectories that better match students of different capacities, and (2) How can we select trajectories that are both informative and well aligned with the student model?

In this work, we propose \textbf{MI-Distillation}, a CoT distillation framework that constructs a continuous Instruct-Reasoning data spectrum through model interpolation. By interpolating between instruction-oriented and reasoning-oriented models, our framework generates reasoning trajectories with varying length, correctness, and reasoning depth. We show that this spectrum changes systematically with the interpolation coefficient $\lambda$, revealing controllable variations in trajectory quality and reasoning length. To select suitable supervision from this spectrum, we introduce \textbf{SeqLSS} (\textbf{Seq}uential \textbf{L}earnable \textbf{S}urprisal \textbf{S}core), a Long CoT selection criterion that balances reasoning information density with student-induced distributional shift. This allows small student models to learn from trajectories that are both informative and well aligned with their intrinsic capabilities.

To summarize, our main contributions are as follows:
\begin{itemize}[leftmargin=*, nosep]
    \item We analyze the gradient dynamics of Long and Short CoT distillation, explaining both the difficulty of Long CoT distillation for small models and the role of model scale in reasoning transfer.
    \item We propose \textbf{MI-Distillation}, which builds a fine-grained Instruct-Reasoning data spectrum via model interpolation, and introduce \textbf{SeqLSS} for selecting student-aligned reasoning trajectories.
    \item Extensive experiments on challenging reasoning benchmarks show that MI-Distillation consistently improves small model reasoning and surpasses strong CoT distillation baselines.
\end{itemize}

\begin{table*}[t]
\centering
\footnotesize
\setlength{\tabcolsep}{3.4pt}
\renewcommand{\arraystretch}{1.15}

% --- Heatmap palette: 4 tiers each (positive=red, negative=blue) ---
\definecolor{posA}{HTML}{FDF3F3}\definecolor{posB}{HTML}{FAE2E2}
\definecolor{posC}{HTML}{F4CACA}\definecolor{posD}{HTML}{EBADAD}
\definecolor{negA}{HTML}{F3F3FD}\definecolor{negB}{HTML}{E2E2FA}
\definecolor{negC}{HTML}{CACAF4}\definecolor{negD}{HTML}{ADADEB}

\newcommand{\up}[2]{\cellcolor{#1}$\uparrow\,#2$}
\newcommand{\dn}[2]{\cellcolor{#1}$\downarrow\,#2$}
\newcommand{\sw}[1]{{\setlength{\fboxsep}{0pt}\colorbox{#1}{\rule{0pt}{1.3ex}\rule{1.3ex}{0pt}}}}

\begin{tabular}{@{}c cccc cccc cccc cccc@{}}
\toprule
\multirow{2}{*}{\textbf{Scale}}
& \multicolumn{8}{c}{\textbf{Nuclear Norm}}
& \multicolumn{8}{c}{\textbf{Effective Rank}} \\
\cmidrule(lr){2-9}\cmidrule(lr){10-17}
& Proj & Short & Long & $\Delta$ & Proj & Short & Long & $\Delta$
& Proj & Short & Long & $\Delta$ & Proj & Short & Long & $\Delta$ \\
\midrule
\multirow{2}{*}{\textbf{1.5B}}
& K & 0.88 & 0.91 & \up{posA}{0.04} & Q & 0.89 & 1.09 & \up{posA}{0.20}
& K & 5.58 & 5.86  & \up{posA}{0.28} & Q &  9.84 & 10.16 & \up{posA}{0.32} \\
& V & 1.60 & 1.67 & \up{posA}{0.08} & O & 1.80 & 2.22 & \up{posB}{0.42}
& V & 8.97 & 7.00  & \dn{negB}{1.97} & O & 13.03 & 12.70 & \dn{negA}{0.33} \\
\midrule
\multirow{2}{*}{\textbf{3B}}
& K & 0.84 & 1.03 & \up{posA}{0.19} & Q & 0.94 & 1.32 & \up{posB}{0.38}
& K & 6.60 & 6.31  & \dn{negA}{0.29} & Q & 10.74 &  8.42 & \dn{negB}{2.32} \\
& V & 1.59 & 2.07 & \up{posB}{0.47} & O & 1.79 & 2.58 & \up{posC}{0.79}
& V & 9.76 & 5.60  & \dn{negC}{4.17} & O & 14.86 &  8.78 & \dn{negD}{6.08} \\
\midrule
\multirow{2}{*}{\textbf{7B}}
& K & 1.12 & 1.76 & \up{posC}{0.65} & Q & 1.11 & 2.00 & \up{posC}{0.88}
& K & 9.74  & 7.76 & \dn{negB}{1.98} & Q & 13.18 & 10.96 & \dn{negB}{2.22} \\
& V & 2.08 & 3.36 & \up{posD}{1.28} & O & 2.05 & 3.68 & \up{posD}{1.63}
& V & 11.09 & 6.36 & \dn{negC}{4.73} & O & 15.94 & 10.34 & \dn{negD}{5.60} \\
\bottomrule
\end{tabular}
\caption{%
\textbf{Attention-projection gradient spectra under Short vs.\ Long CoT distillation.}
$\Delta = \text{Long}-\text{Short}$ (denoted by $\uparrow/\downarrow$); each entry is averaged over 500 MATH-500 examples and all attention layers.
Cell shading encodes $|\Delta|$ in four bins:
\sw{posA}\,\sw{posB}\,\sw{posC}\,\sw{posD}\, for nuclear-norm growth and
\sw{negA}\,\sw{negB}\,\sw{negC}\,\sw{negD}\, for effective-rank collapse.
Long CoT uniformly inflates nuclear norms with gaps widening as the student scales up, while effective-rank reductions become pronounced from 3B onward.%
}
\label{tab:attn-spectral-gap}
\vspace{-1em}
\end{table*}

\section{Related Work}
\subsection{Reasoning in Large Language Models}
Chain-of-thought (CoT) reasoning has become a key mechanism for enhancing the reasoning capabilities of large language models~\cite{wangr4,jaech2024openai}. Early distillation paradigms typically rely on Short CoTs, either manually annotated or generated by instruction-tuned models, to provide concise intermediate reasoning steps. More recently, reinforcement learning with verifiable rewards (RLVR) ~\cite{shao2024deepseekmath,yu2025dapo} has enabled language models to produce Long CoTs with detailed reflection, and verification, substantially improving performance on complex mathematical and symbolic reasoning tasks~\cite{he2024olympiadbench,xu2024faithful}.

However, Long CoT trajectories are not universally beneficial as distillation targets. Their rich reasoning patterns may also increase imitation difficulty for smaller students, particularly when the teacher rationale is poorly aligned with the student's learning capacity and intrinsic distribution. This motivates reasoning supervision that balances informativeness with student-side learnability.

\subsection{Knowledge Distillation}
Knowledge distillation~\cite{hinton2015distilling} transfers the capabilities of large language models to smaller students by training them on outputs from stronger teachers. However, recent studies suggest that directly distilling Long CoT trajectories into small models does not consistently improve performance, as small students often benefit more from shorter and more concise rationales~\cite{li2025small,kim2025their,jiang2025teach}. 

Existing approaches address this challenge by mixing short and long CoT data or introducing curriculum-based distillation, gradually increasing the complexity of the supervision~\cite{jiang2025teach,li2025small}. Despite their effectiveness, these methods largely depend on manually designed data mixtures or schedules, and offer limited control over which reasoning traces are most suitable for a given student model. Our work instead analyzes the scale-dependent effects of short and long CoT distillation, and constructs a continuous reasoning-data spectrum via model interpolation, enabling more adaptive selection of CoT trajectories for small-model distillation.

\begin{figure*}[t]
    \centering
    \includegraphics[width=\linewidth]{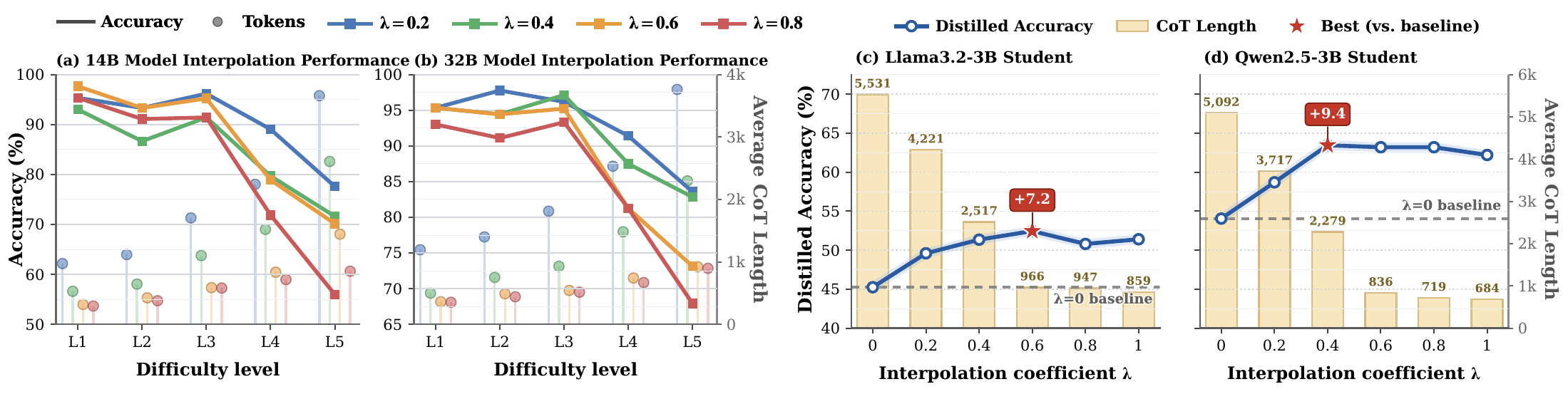}
    \caption{
         \textbf{Effect of interpolation coefficient $\lambda$ on teacher performance and CoT distillation on MATH-500}.
    The left panel evaluates interpolated teachers across difficulty levels, where varying $\lambda$ smoothly shifts the teacher from reasoning-oriented to instruction-oriented behavior. The right panel reports the final accuracy of Llama3.2-3B and Qwen2.5-3B students distilled from CoT trajectories generated by 32B interpolated teacher. Intermediate teachers achieve a better balance between reasoning depth and student learnability, leading to stronger distillation performance for small models.
    }
    \label{fig:interpolation_analysis}
    \vspace{-8pt}
\end{figure*}

\section{Gradient Signatures of Short and Long CoT Distillation}
\label{sec:gradient_dynamics}
While recent studies indicate that smaller models benefit more from Short CoT distillation than Long CoT, the optimization behavior underlying this phenomenon remains poorly understood. To bridge this gap, we analyze the gradient signatures induced by SFT across student model sizes.

Adopting the spectral analysis from \citet{li2024happened}, we characterize the gradients of the attention projection matrices (\textit{Query}, \textit{Key}, \textit{Value}, and \textit{Output}) via Singular Value Decomposition (SVD) \citep{carlini2024stealing}. For a given gradient matrix $G \in \mathbb{R}^{m \times n}$, we summarize its magnitude and structural concentration using the \textbf{Nuclear Norm} ($\|G\|_*$) and \textbf{Effective Rank} ($\operatorname{ERank}$):
\begin{align}
    \|G\|_* &= \sum_{k=1}^{\min(m,n)} \sigma_k, \label{eq:nuclear_norm} \\
    \operatorname{ERank}(G) &= \exp\!\left( -\sum_{k=1}^{\min(m,n)} p_k\log p_k \right), \label{eq:erank}
\end{align}
where $\sigma_k$ represents the $k$-th singular value of $G$, and $p_k = \sigma_k / \|G\|_*$ defines the normalized singular-value distribution.

From an optimization perspective, the nuclear norm captures the total spectral energy of the gradients; a higher value signifies a more substantial optimization signal imparted by the distillation process. Conversely, the effective rank acts as a continuous measure of dimensionality by computing the Shannon entropy of the gradient spectrum. A lower effective rank indicates that the parameter updates are structurally concentrated along a few dominant principal directions, leading to lower-dimensional optimization trajectories. Detailed layer-wise formulations are provided in Appendix~\ref{sec:appendix_gradient}.

\subsection{Results and Analysis}
We generate Short and Long CoT trajectories on MATH-500 using Qwen2.5-32B-Instruct~\cite{qwen2.5} and QwQ-32B~\cite{qwq32b}, respectively, and analyze gradients from Qwen2.5 students at 1.5B, 3B, and 7B scales. Dataset details are provided in Appendix~\ref{evaluation_datasets}, with results reported in Table~\ref{tab:attn-spectral-gap}.

Across all model scales, Long CoT consistently induces larger nuclear norms than Short CoT, indicating stronger optimization signals during distillation. This suggests that Long CoT supervision carries richer gradient information, but also imposes a more demanding learning target for the student. Notably, the nuclear-norm gap becomes more pronounced as the student scale increases, especially for the 3B and 7B models. This trend indicates that larger students are better able to absorb the additional supervision provided by long reasoning trajectories and translate it into stronger spectral gradient updates. Effective rank reveals a complementary trend. From the 3B scale onward, Long CoT produces lower effective rank than Short CoT, showing that its gradients become more concentrated along dominant spectral directions. Thus, for sufficiently capable students, Long CoT induces updates that are not only larger in magnitude but also more structurally coherent.

% The effective-rank results further reveal a scale-dependent structural difference. From the 3B model onward, Long CoT yields lower effective ranks than Short CoT, showing that its stronger gradients are also more concentrated along a smaller number of dominant spectral directions. In other words, larger students can transform long reasoning trajectories into more coherent and low-dimensional update directions. By contrast, for smaller students, Long CoT may behave more like a high-loss imitation target than an effective source of transferable reasoning signal.

This motivates a central principle of our method: \textbf{Effective Long CoT distillation should balance the information density of teacher reasoning trajectories with the distributional and optimization capacity of the student model}.

\section{Synthesizing Instruct-Reasoning Spectrum via Model Interpolation}
\subsection{Preliminaries} 
Motivated by our gradient analysis, we find that small students are often poorly aligned with the complex System2 style trajectories produced by LRMs, whereas System1 style Short CoT offers more learnable but less informative supervision. Existing methods typically rely on discrete mixtures of Short and Long CoT data, which offer limited granularity in controlling the reasoning signal. Building on recent advances in model merging and interpolation~\cite{wu2025revisiting}, we interpolate instruction-oriented and reasoning-oriented teachers to construct a family of intermediate models with smoothly varying reasoning depth and reasoning compactness behavior. The resulting Instruct-Reasoning spectrum provides diverse CoT trajectories between Short and Long reasoning, allowing students to receive supervision that better balances reasoning richness and capacity-aware learnability.

Let $\Theta^{\mathrm{Thi}}$ and $\Theta^{\mathrm{Ins}}$ denote
the reasoning-oriented and instruction-oriented teachers.
We synthesize intermediate teachers through parameter interpolation,
\begin{equation}
\Theta^{\mathrm{MI}}_{\lambda}
\;=\;
\lambda\, \Theta^{\mathrm{Ins}}
+ (1-\lambda)\, \Theta^{\mathrm{Thi}},
\quad \lambda \in [0,1].
\label{eq:mi-interp}
\end{equation}
This operation is independent of the choice of reference model
and can be interpreted as a special case of task arithmetic~\cite{ilharco2022editing}.
For any base model $\Theta^{\mathrm{Base}}$, define the task vectors
$\bm{\tau}^{\mathrm{Thi}} = \Theta^{\mathrm{Thi}} - \Theta^{\mathrm{Base}}$
and
$\bm{\tau}^{\mathrm{Ins}} = \Theta^{\mathrm{Ins}} - \Theta^{\mathrm{Base}}$.
Substituting these definitions into Eq.~\eqref{eq:mi-interp} yields
\begin{align}
\Theta^{\mathrm{MI}}_{\lambda}
&= \lambda\, \Theta^{\mathrm{Ins}}
   + (1-\lambda)\, \Theta^{\mathrm{Thi}} \notag \\
&= \lambda \bigl( \bm{\tau}^{\mathrm{Ins}} + \Theta^{\mathrm{Base}} \bigr)
   + (1-\lambda) \bigl( \bm{\tau}^{\mathrm{Thi}} + \Theta^{\mathrm{Base}} \bigr)
   \notag \\
&= \Theta^{\mathrm{Base}}
   + \lambda\, \bm{\tau}^{\mathrm{Ins}}
   + (1-\lambda)\, \bm{\tau}^{\mathrm{Thi}}.
\label{eq:mi-taskarith}
\end{align}
Therefore, model interpolation can be viewed as task arithmetic
over reasoning and instruction task vectors. Varying $\lambda$ yields a continuous teacher spectrum that
smoothly balances reasoning compactness behavior and reasoning depth.

\subsection{Evaluating Interpolated Teachers and Distillation Effectiveness}
To further examine the effectiveness of the interpolated teachers, we evaluate models obtained by interpolating two model families: the Qwen-14B family, including DeepSeek-R1-Distill-14B and Qwen2.5-14B-Instruct, and the Qwen-32B family, including QwQ-32B~\cite{qwq32b} and Qwen2.5-32B-Instruct~\cite{qwen2.5}. We first assess these interpolated teachers on MATH-500~\cite{hendrycks2021measuring} across different difficulty levels, which allows us to characterize how model interpolation modulates reasoning capability and problem-solving behavior. 

We then assess the downstream utility of this interpolation-induced spectrum by distilling CoT trajectories generated by all 32B interpolated teachers into smaller students, including Qwen2.5-3B-Instruct and Llama3.2-3B-Instruct~\cite{grattafiori2024llama}. Notably, the two 32B endpoint teachers are variants within the Qwen2.5-32B family, which largely controls for differences in world knowledge and pretraining background. This setting allows us to more directly attribute downstream distillation effects to interpolation-induced changes in reasoning style, offering a cleaner evaluation of the proposed Instruct-Reasoning spectrum.

\paragraph{Results and Analysis.}
Figures~\ref{fig:interpolation_analysis}(a) and \ref{fig:interpolation_analysis}(b) present the evaluation results of interpolated teacher models at the 14B and 32B scales. Consistent with our preceding analysis, varying the interpolation coefficient induces a clear trade-off between reasoning correctness and reasoning depth. The resulting teachers form an approximate spectrum of reasoning behaviors, with CoT length changing smoothly across interpolation settings. This trend indicates that model interpolation successfully produces diverse reasoning trajectories beyond the two endpoint teachers.

Figures \ref{fig:interpolation_analysis}(c) and \ref{fig:interpolation_analysis}(d) further present the distillation results on Qwen and Llama student families using CoT data generated by 32B teachers at different interpolation coefficients. The results provide additional evidence for our hypothesis: the reasoning length of the distilled students also varies smoothly with the interpolation coefficient, indicating that the Instruct-Reasoning spectrum is effectively transferred through distillation. More interestingly, the best student performance is typically achieved not at either endpoint, i.e., $\lambda=0$ corresponding to the reasoning-oriented teacher or $\lambda=1$ corresponding to the instruction-oriented teacher, but at an intermediate interpolation point. This observation supports our central claim that effective CoT distillation does not solely depend on the informativeness of reasoning trajectories or their alignment with the student distribution. Instead, the most beneficial supervision often emerges from a balanced trade-off between reasoning information density and student-side learnability.
Complete evaluation results of both interpolated teachers and distilled students on additional benchmarks are provided in Appendix~\ref{sec:extended_results}. In addition, Appendix~\ref{app:ties_comparison} compares direct parameter interpolation with TIES-Merging~\cite{yadav2023ties}, showing that the resulting teacher spectrum is robust to the choice of merging strategy.

\begin{figure*}[t]
    \centering
    \includegraphics[width=\linewidth]{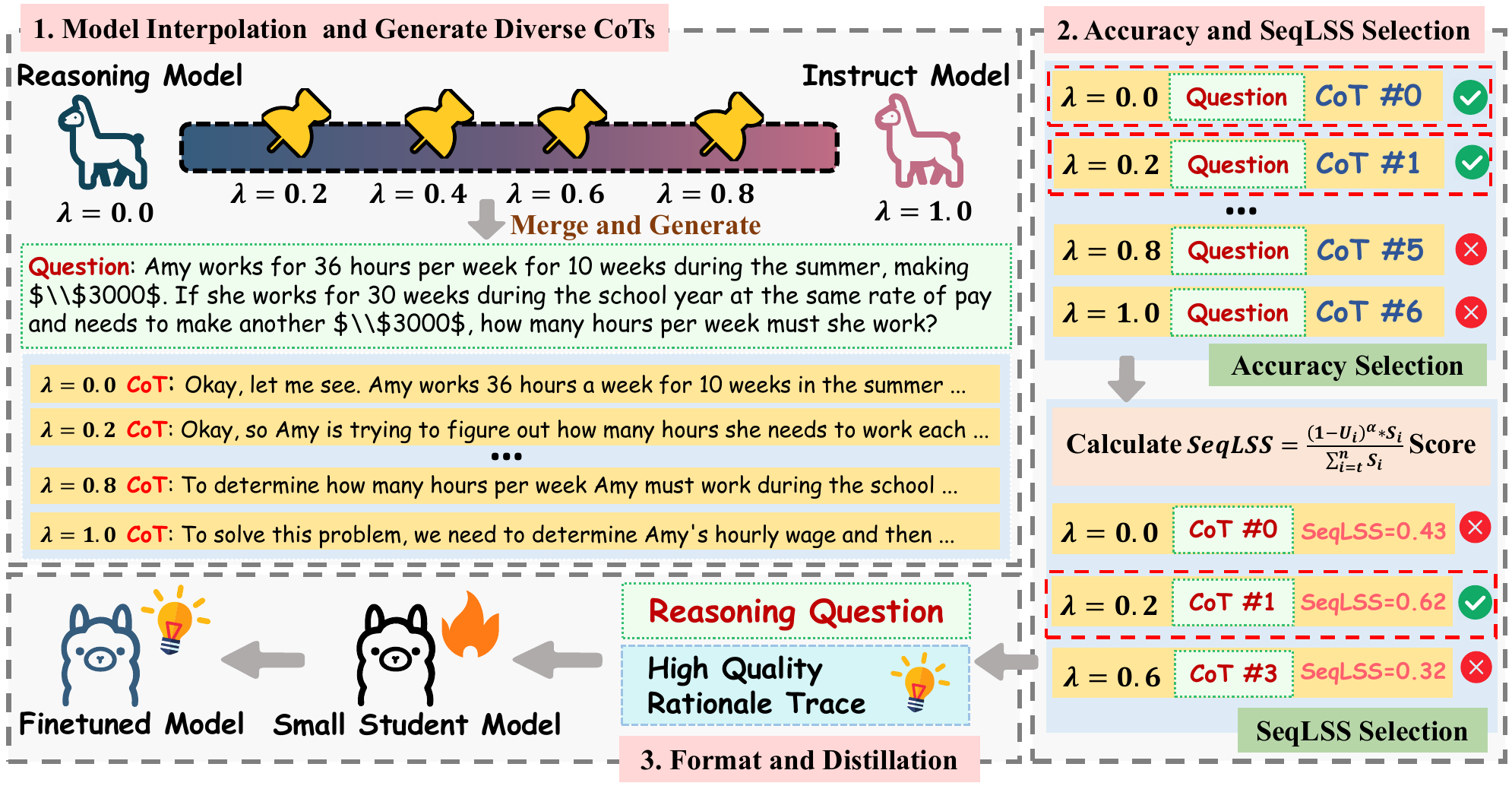}
    \caption{
\textbf{Overview of the proposed MI-Distillation framework.} 
MI-Distillation first constructs an instruct-reasoning spectrum via model interpolation and generates diverse CoT trajectories from the interpolated teachers. 
It then filters the candidates by answer correctness and selects student-aligned rationales using the proposed \textbf{SeqLSS} criterion. 
The selected question--rationale pairs are finally used to distill a small student model.
}
    \label{fig:mi_distillation_pipeline}
    \vspace{-1em}
\end{figure*}

% Add to preamble if not already included:
% \usepackage[table]{xcolor}

\begin{table*}[t]
\centering
\resizebox{\textwidth}{!}{
\begin{tabular}{l ccc ccc cc}
\toprule
\multirow{2}{*}{\textbf{Distillation Method}} 
& \multicolumn{7}{c}{\textbf{Evaluation Datasets}} 
& \multirow{2}{*}{\textbf{Avg.}} \\
\cmidrule(lr){2-8}
& \textbf{AIME24} 
& \textbf{AMC23} 
& \textbf{GPQA-Diamond} 
& \textbf{GSM8K} 
& \textbf{MATH-500} 
& \textbf{Minerva} 
& \textbf{Olympiad} 
& \\
\midrule

\rowcolor{gray!10}
\multicolumn{9}{c}{\textit{\textbf{Student Model: Qwen2.5-3B-Instruct}}} \\
\midrule

\method{Short CoT (32B-Instruct)} 
& $\mathbf{5.42}_{\pm 1.67}$ 
& $35.00_{\pm 6.65}$ 
& $25.88_{\pm 2.87}$ 
& $83.49_{\pm 1.04}$ 
& $62.20_{\pm 1.97}$ 
& $25.37_{\pm 1.59}$ 
& $27.00_{\pm 0.94}$ 
& 37.77 \\

\method{Long CoT (QwQ-32B)} 
& $3.75_{\pm 3.19}$ 
& $27.03_{\pm 4.10}$ 
& $19.32_{\pm 0.96}$ 
& $80.78_{\pm 0.43}$ 
& $54.05_{\pm 0.72}$ 
& $20.13_{\pm 1.25}$ 
& $21.11_{\pm 1.18}$ 
& 32.31 \\

\method{Mix Large} 
& $\underline{5.21}_{\pm 2.42}$ 
& $\underline{36.09}_{\pm 6.45}$ 
& $\underline{29.29}_{\pm 2.10}$ 
& $\underline{84.74}_{\pm 0.57}$ 
& $\underline{63.00}_{\pm 0.98}$ 
& $\mathbf{26.38}_{\pm 1.42}$ 
& $\underline{28.89}_{\pm 1.36}$ 
& \underline{39.09} \\

\method{Mix Long} 
& $2.29_{\pm 2.35}$ 
& $30.63_{\pm 4.96}$ 
& $22.60_{\pm 2.20}$ 
& $83.00_{\pm 1.04}$ 
& $57.40_{\pm 1.93}$ 
& $22.33_{\pm 2.11}$ 
& $23.96_{\pm 1.03}$ 
& 34.60 \\

\method{Curriculum Distillation} 
& $3.54_{\pm 3.54}$ 
& $29.38_{\pm 4.70}$ 
& $24.24_{\pm 4.38}$ 
& $81.22_{\pm 1.00}$ 
& $56.15_{\pm 1.11}$ 
& $21.51_{\pm 2.18}$ 
& $22.85_{\pm 1.05}$ 
& 34.13 \\

\rowcolor{hlblue}
\textbf{\method{MI-Distillation} (Ours)} 
& $\mathbf{5.42}_{\pm 2.06}$ 
& $\mathbf{40.31}_{\pm 5.39}$ 
& $\mathbf{29.55}_{\pm 1.27}$ 
& $\mathbf{85.06}_{\pm 0.53}$ 
& $\mathbf{66.15}_{\pm 1.08}$ 
& $\underline{25.46}_{\pm 1.10}$ 
& $\mathbf{29.49}_{\pm 0.81}$ 
& \textbf{40.21} \\

\midrule

\rowcolor{gray!10}
\multicolumn{9}{c}{\textit{\textbf{Student Model: Llama3.2-3B-Instruct}}} \\
\midrule

\method{Short CoT (32B-Instruct)} 
& $3.96_{\pm 2.78}$ 
& $24.38_{\pm 5.88}$ 
& $23.99_{\pm 2.35}$ 
& $76.65_{\pm 1.17}$ 
& $51.40_{\pm 1.25}$ 
& $\underline{17.10}_{\pm 2.43}$ 
& $17.43_{\pm 1.44}$ 
& 30.70 \\

\method{Long CoT (QwQ-32B)} 
& $0.83_{\pm 1.49}$ 
& $20.00_{\pm 4.18}$ 
& $22.47_{\pm 2.69}$ 
& $\mathbf{78.53}_{\pm 0.97}$ 
& $45.25_{\pm 2.46}$ 
& $13.24_{\pm 1.08}$ 
& $13.87_{\pm 1.00}$ 
& 27.74 \\

\method{Mix Large} 
& $\underline{4.58}_{\pm 2.95}$ 
& $\underline{27.19}_{\pm 4.82}$ 
& $24.37_{\pm 1.99}$ 
& $76.88_{\pm 0.45}$ 
& $\underline{52.90}_{\pm 1.18}$ 
& $\mathbf{17.19}_{\pm 0.63}$ 
& $\underline{17.58}_{\pm 0.82}$ 
& \underline{31.53} \\

\method{Mix Long} 
& $1.67_{\pm 2.11}$ 
& $21.09_{\pm 4.91}$ 
& $\mathbf{26.26}_{\pm 2.37}$ 
& $\underline{77.84}_{\pm 0.99}$ 
& $44.55_{\pm 1.58}$ 
& $14.43_{\pm 1.48}$ 
& $14.43_{\pm 0.31}$ 
& 28.61 \\

\method{Curriculum Distillation} 
& $0.42_{\pm 1.14}$ 
& $19.53_{\pm 5.72}$ 
& $21.09_{\pm 2.55}$ 
& $73.41_{\pm 0.59}$ 
& $44.00_{\pm 0.85}$ 
& $12.41_{\pm 2.67}$ 
& $14.73_{\pm 0.85}$ 
& 26.51 \\

\rowcolor{hlblue}
\textbf{\method{MI-Distillation} (Ours)} 
& $\mathbf{6.88}_{\pm 3.10}$ 
& $\mathbf{30.47}_{\pm 2.28}$ 
& $\underline{24.62}_{\pm 2.52}$ 
& $77.75_{\pm 0.77}$ 
& $\mathbf{54.60}_{\pm 1.02}$ 
& $16.27_{\pm 0.46}$ 
& $\mathbf{19.96}_{\pm 0.28}$ 
& \textbf{32.93} \\

\bottomrule
\end{tabular}
}
\caption{\textbf{Performance comparison of different distillation methods on Qwen2.5-3B-Instruct and Llama3.2-3B-Instruct student models.} 
For each student-model group, the best result is highlighted in \textbf{bold}, while the second-best result is \underline{underlined}. 
Standard deviations are reported as subscripts for compactness.}
\label{tab:main_results}
\end{table*}

\begin{table*}[t]
\centering
\small
\setlength{\tabcolsep}{8pt}
\renewcommand{\arraystretch}{1.15}
\resizebox{\textwidth}{!}{
\begin{tabular}{l ccccc c}
\toprule
\multirow{2}{*}{\textbf{Selection Method}}
& \multicolumn{5}{c}{\textbf{Evaluation Datasets}}
& \multirow{2}{*}{\textbf{Avg.}} \\
\cmidrule(lr){2-6}
& \textbf{GPQA-Diamond}
& \textbf{GSM8K}
& \textbf{MATH-500}
& \textbf{Minerva}
& \textbf{Olympiad}
& \\
\midrule
\method{PPL}~\cite{marion2023less}
& $\underline{27.15}_{\pm 4.25}$
& $\mathbf{85.42}_{\pm 0.59}$
& $\underline{64.50}_{\pm 1.18}$
& $18.29_{\pm 2.04}$
& $\underline{29.12}_{\pm 1.06}$
& $\underline{44.90}$ \\
\method{IFD}~\cite{li2024quantity}
& $26.89_{\pm 2.98}$
& $84.69_{\pm 0.82}$
& $63.70_{\pm 1.83}$
& $17.46_{\pm 1.10}$
& $26.34_{\pm 1.23}$
& $43.82$ \\
\method{Random}
& $26.01_{\pm 3.97}$
& $84.48_{\pm 0.33}$
& $63.25_{\pm 1.34}$
& $\underline{19.21}_{\pm 1.54}$
& $26.78_{\pm 0.77}$
& $43.95$ \\
\method{SeqLSS-min}
& $26.77_{\pm 1.89}$
& $83.55_{\pm 0.21}$
& $59.00_{\pm 1.82}$
& $16.36_{\pm 1.66}$
& $24.81_{\pm 1.54}$
& $42.10$ \\
\rowcolor{hlblue}
\textbf{\method{SeqLSS} (Ours)}
& $\mathbf{29.55}_{\pm 1.27}$
& $\underline{85.06}_{\pm 0.53}$
& $\mathbf{66.15}_{\pm 1.08}$
& $\mathbf{25.46}_{\pm 1.10}$
& $\mathbf{29.49}_{\pm 0.81}$
& $\mathbf{47.14}$ \\
\bottomrule
\end{tabular}
}
\caption{
\textbf{Ablation on the SeqLSS selection criterion with Qwen2.5-3B-Instruct as the student model.}
All methods select trajectories from the identical candidate pool generated by the interpolated teachers under the same answer-correctness filtering, and differ only in the ranking criterion.
SeqLSS-min selects the trajectories with the \emph{lowest} SeqLSS scores.
The best result in each column is shown in \textbf{bold}, and the second-best is \underline{underlined}.
Standard deviations are reported as subscripts for compactness.
}
\label{tab:seqlss_ablation}
\vspace{-1em}
\end{table*}

\section{MI-Distillation: SeqLSS-Guided CoT Selection from Interpolated Teachers}
Building on the interpolated teacher spectrum, we propose \textbf{M}odel-\textbf{I}nterpolation \textbf{Distillation} (\textbf{MI-Distillation}), a student-aware framework for small model CoT distillation. MI-Distillation leverages interpolated teachers to construct a fine-grained spectrum of CoT trajectories, rather than relying solely on instruction-oriented or reasoning-oriented endpoints. Since a fixed interpolation coefficient may not suit students with different capacities or model families, we further introduce \textbf{Seq}uential \textbf{L}earnable \textbf{S}urprisal \textbf{S}core (\textbf{SeqLSS}) to select trajectories that balance reasoning informativeness with student learnability. This enables MI-Distillation to provide supervision that is both informative and aligned with the student’s learning capacity, improving the efficiency and effectiveness of CoT distillation for small models.

\subsection{Model Interpolation and Generate Diverse CoTs}
Following the model interpolation defined in Eq.~\eqref{eq:mi-interp}, the first stage instantiates a set of interpolated teachers from a reasoning-oriented model and an instruction-oriented model. Specifically, we vary the interpolation coefficient over
\(\Lambda=\{0.2,0.4,0.6,0.8,1.0\}\), yielding a family of teacher models \(\{M_{\lambda}\}_{\lambda\in\Lambda}\) with parameters \(\{\Theta_{\lambda}\}_{\lambda\in\Lambda}\). These teachers exhibit different trade-offs between reasoning depth, response compactness, and instruction-oriented behavior.

Given a training set \(\mathcal{S}=\{(q_i,a_i)\}_{i=1}^{N}\), each interpolated teacher \(M_{\lambda}\) is prompted to generate a CoT trajectory for question \(q_i\):
\begin{equation}
\mathbf{R}_{i,\lambda} \sim M_{\lambda}(\cdot \mid q_i),
\quad \lambda \in \Lambda .
\end{equation}
The generated trajectories naturally differ in correctness, complexity, and reasoning granularity. We collect them into an Instruct-Reasoning spectrum:
\begin{equation}
\mathcal{D}
=
\{(q_i,a_i,\lambda,\mathbf{R}_{i,\lambda})
\mid
(q_i,a_i)\in\mathcal{S},\lambda\in\Lambda\}.
\end{equation}
Rather than committing to a single teacher or a fixed Short/Long CoT style, \(\mathcal{D}\) exposes a structured set of candidate trajectories along the instruct-reasoning spectrum. This design allows MI-Distillation to perform student-aware CoT selection from a richer supervision space, instead of directly distilling from either endpoint teacher.

\subsection{Accuracy and SeqLSS Selection}
\paragraph{Accuracy Selection.}
We first filter candidate trajectories by answer correctness, since erroneous CoTs can introduce misleading supervision and bias the student toward invalid reasoning patterns~\cite{xiang2025towards}. 
For each trajectory $\mathbf{R}_{i,\lambda}$ generated along the Instruct-Reasoning spectrum, we retain it only when its final prediction matches the ground-truth answer $a_i$.

\paragraph{SeqLSS Selection.}
After correctness filtering, multiple valid rationales may still differ substantially in how useful they are for a specific student model. An ideal CoT trajectory should be informative enough to provide non-trivial reasoning signals, while remaining aligned with the student's current predictive distribution. To capture this trade-off, we introduce a student-conditioned criterion, \textbf{L}earnable \textbf{S}urprisal \textbf{S}core (\textbf{LSS}), defined at the token level.
Given an input problem $x$ and a candidate rationale $R = (r_1, \ldots, r_T)$, the \emph{surprisal} of token $r_i$ under the student model $\theta_s$ is
\begin{equation}
    S_i \;=\; -\log p_{\theta_s}(r_i \mid x, r_{<i}),
\end{equation}
which quantifies how much new information $r_i$ conveys to the student. However, high-surprisal tokens are not always beneficial: they may lie far outside the student's learnable region. We measure this distributional misalignment by the cumulative probability mass that the student assigns to tokens ranked above $r_i$:
\begin{equation}
    U_i \;=\!\!\!\! \sum_{\substack{v \,\in\, \mathcal{V} \\ p_{\theta_s}(v \mid x, r_{<i}) \,>\, p_{\theta_s}(r_i \mid x, r_{<i})}} \!\!\!\! p_{\theta_s}(v \mid x, r_{<i}).
\end{equation}
Intuitively, $U_i$ reflects how much probability mass the student assigns to tokens ranked above the target token; a smaller $U_i$ indicates that $r_i$ lies closer to the student's high-probability region and is therefore easier to learn.
The token-level LSS combines these two quantities:
\begin{equation}
    \mathrm{LSS}_i \;=\; S_i \, (1 - U_i)^{\alpha},
\end{equation}
where $S_i$ rewards informative tokens and $(1 - U_i)^{\alpha}$ down-weights tokens that fall outside the student's high-probability region. The exponent $\alpha \geq 0$ controls the strength of this learnability penalty, so that $\mathrm{LSS}_i$ favors tokens that are simultaneously informative and learnable.
Naively averaging $\mathrm{LSS}_i$ over a rationale is unstable, since a few tokens with extreme surprisal may dominate the sequence-level score~\cite{yang2026reasoning}. We therefore aggregate token-level scores using surprisal as a normalization signal, yielding the \textbf{Sequential Learnable Surprisal Score}:
\begin{equation}
    \mathrm{SeqLSS}(\mathbf{R}) \;=\; \frac{\sum_{i=1}^{T} S_i \, (1 - U_i)^{\alpha}}{\sum_{i=1}^{T} S_i}.
\end{equation}
Intuitively, $\mathrm{SeqLSS}(\mathbf{R})$ estimates the proportion of informative reasoning signal in $\mathbf{R}$ that lies within the student's learnable region. For each problem, we select the correct candidate rationale with the highest $\mathrm{SeqLSS}$, thereby balancing reasoning-information density with student-side distributional alignment for CoT distillation.

\paragraph{Format and Distillation.}
After accuracy and SeqLSS-based selection, the retained trajectories form the final distillation set
$\mathcal{D}_{\mathrm{select}}=\{(q_i,R_i)\}$, where $q_i$ denotes the input question and $R_i$ is the selected response trajectory, including both the reasoning process and the final answer.
The student model $\theta_S$ is trained on $\mathcal{D}_{\mathrm{select}}$ with the standard next-token prediction objective:
\begin{equation}
\mathcal{L}(\theta_S)
=
-\sum_{(q_i,R_i)\in \mathcal{D}_{\mathrm{select}}}
\log p_{\theta_S}(R_i \mid q_i).
\end{equation}
The resulting student model is then evaluated on downstream reasoning benchmarks.

\section{Experiments}
\paragraph{Experiment Details.}
To evaluate the effectiveness and robustness of MI-Distillation, we conduct experiments on two student models from different model families: Qwen2.5-3B-Instruct and Llama3.2-3B-Instruct. 
We use QwQ-32B as the Long CoT teacher and Qwen2.5-32B-Instruct as the Short CoT teacher. 
For data generation, we follow the recommended decoding setting with temperature 0.6, top-$p$ 0.95~\cite{guo2025deepseek}, and a maximum generation length of 8192 tokens. 
The training corpus is constructed from 7,500 randomly sampled problems from the MATH training set~\cite{hendrycks2021measuring}. 
For each problem, we generate candidate reasoning trajectories using the two endpoint teachers as well as the interpolated teachers in our MI-Distillation framework. Unless otherwise specified, the learnability penalty coefficient $\alpha$ is empirically set to 4. 

\paragraph{Training and Evaluation.}
All student models are fine-tuned for 3 epochs using LLaMA-Factory~\cite{zheng2024llamafactory}, with a global batch size of 32 and a peak learning rate of $1\times10^{-5}$. 
To comprehensively assess the distilled students, we evaluate them on a suite of reasoning benchmarks with varying difficulty, including \textbf{GSM8K}~\cite{cobbe2021gsm8k}, \textbf{MATH-500}~\cite{hendrycks2021measuring}, \textbf{AMC23}~\cite{AMC}, \textbf{AIME24}~\cite{aime24}, \textbf{GPQA-Diamond}~\cite{rein2023gpqa}, \textbf{Minerva}~\cite{lewkowycz2022solving}, and \textbf{OlympiadBench}~\cite{he2024olympiadbench}. 
We follow the teacher-generation decoding protocol and report Pass@1 with the corresponding standard deviation.
Further details regarding hardware configurations and inference setups are provided in Appendix~\ref{sec:appendix_exp_details}.

\paragraph{Baselines.}
In addition to standard distillation from Short CoT and Long CoT teachers, we compare MI-Distillation against three stronger baselines that combine or schedule different types of CoT supervision. 
\textbf{Mix Long}~\cite{li2025small} randomly mixes Short-CoT and Long-CoT trajectories at a 4:1 ratio, providing a simple data-level combination of concise and elaborated reasoning traces. 
\textbf{Mix Large}~\cite{li2025small} extends this setting by incorporating trajectories generated from teachers of different scales under the same mixing ratio, thereby increasing teacher diversity. 
\textbf{Curriculum Learning}~\cite{jiang2025teach} introduces Long-CoT supervision progressively during training, encouraging the student to gradually adapt from simpler reasoning patterns to more complex ones.

\subsection{Results and Analysis}
\paragraph{Main Results.}
Table~\ref{tab:main_results} reports the main comparison on two 3B student models. 
MI-Distillation achieves the best average performance on both Qwen2.5-3B-Instruct and Llama3.2-3B-Instruct, outperforming the strongest baseline by 1.12 and 1.40 points, respectively. 
It also yields consistent gains on most challenging reasoning benchmarks, including AMC23, MATH-500, and OlympiadBench. 
Consistent with our analysis, Long CoT distillation often underperforms Short CoT distillation, showing that more detailed rationales are not inherently more learnable for small students.
By selecting trajectories from the Instruct-Reasoning spectrum, MI-Distillation better balances reasoning depth and student learnability, leading to more robust improvements across model families.

\begin{table}[!t]
\centering
\small
\setlength{\tabcolsep}{3.5pt}
\renewcommand{\arraystretch}{1.12}
\resizebox{\columnwidth}{!}{
\begin{tabular}{lccc}
\toprule
\textbf{Coefficient} & \textbf{GSM8K} & \textbf{MATH-500} & \textbf{AMC23} \\
\midrule
$\lambda=0.2$ & $83.07_{\pm 1.13}$ & $58.70_{\pm 1.05}$ & $32.34_{\pm 4.96}$ \\
$\lambda=0.4$ & $84.00_{\pm 0.77}$ & $\underline{63.45}_{\pm 0.62}$ & $36.88_{\pm 4.23}$ \\
$\lambda=0.6$ & $84.48_{\pm 0.23}$ & $62.20_{\pm 1.10}$ & $\underline{39.22}_{\pm 6.10}$ \\
$\lambda=0.8$ & $\underline{84.63}_{\pm 0.83}$ & $63.20_{\pm 1.26}$ & $37.97_{\pm 5.72}$ \\
\midrule
\rowcolor{hlblue}
\textbf{MI-Distillation} 
& $\mathbf{84.82}_{\pm 0.66}$ 
& $\mathbf{66.15}_{\pm 1.08}$ 
& $\mathbf{40.31}_{\pm 5.39}$ \\
\bottomrule
\end{tabular}
}
\caption{
Comparison between MI-Distillation and distillation from fixed interpolation coefficients using Qwen2.5-3B-Instruct as the student model.
MI-Distillation consistently outperforms all fixed-$\lambda$ variants across GSM8K, MATH-500, and AMC23, demonstrating the advantage of adaptive trajectory selection over relying on a single interpolated teacher distribution.
}
\label{tab:fixed-lambda-comparison}
\vspace{-8pt}
\end{table}

\paragraph{Comparison with Fixed Interpolation Coefficients.}
Table~\ref{tab:fixed-lambda-comparison} compares MI-Distillation with direct distillation from CoT trajectories generated under fixed interpolation coefficients. 
MI-Distillation consistently achieves the best performance across GSM8K, MATH-500, and AMC23, outperforming the strongest fixed-$\lambda$ baseline by 0.19, 2.70, and 1.09 points, respectively. 
Crucially, no single interpolation coefficient dominates across all benchmarks: $\lambda=0.8$ performs best on GSM8K, whereas $\lambda=0.4$ and $\lambda=0.6$ are stronger on MATH-500 and AMC23. 
This highlights the limitation of using a globally fixed interpolation coefficient for reasoning distillation. 
In contrast, MI-Distillation adaptively selects trajectories from the Instruct-Reasoning spectrum, enabling the student to benefit from reasoning traces that better balance information density with learnability.

\paragraph{Ablation on the SeqLSS Selection Criterion.}
To verify that the gains of MI-Distillation stem from the proposed ranking criterion itself, we compare SeqLSS against widely used data-selection baselines on the same candidate trajectories generated by model interpolation, including perplexity-based selection (PPL)~\cite{marion2023less}, IFD~\cite{li2024quantity}, and random selection.
We additionally include \textbf{SeqLSS-min}, which instead selects the trajectories with the lowest SeqLSS scores, to examine whether the proposed ranking identifies genuinely useful supervision.
As shown in Table~\ref{tab:seqlss_ablation}, SeqLSS achieves the best average performance and outperforms all baselines on four of the five benchmarks.
In contrast, SeqLSS-min performs worst overall, confirming that the improvement is attributable to the proposed ranking criterion rather than merely to subsampling the candidate pool.
A complementary sensitivity analysis of the learnability penalty coefficient $\alpha$ is provided in Appendix~\ref{app:alpha_ablation}.

\begin{figure}[t]
    \centering
    \includegraphics[width=\columnwidth]{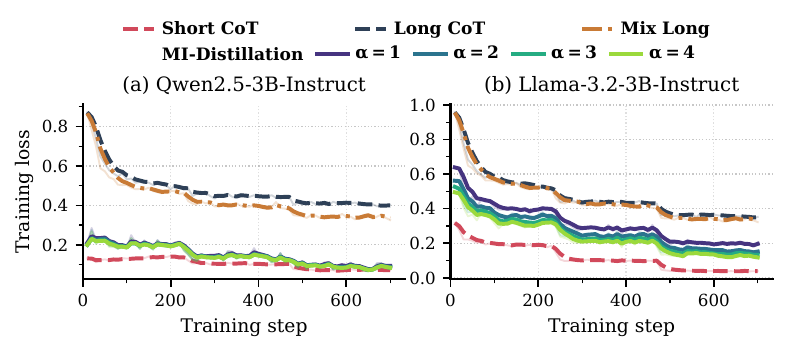}
    \caption{
    \textbf{Training loss of student models under different learnability penalty coefficient $\alpha$}.
    Compared with Long CoT and Mix Long, MI-Distillation exhibits more stable optimization across both student backbones, indicating that learnability-aware selection improves the alignment between reasoning trajectories and student-model capacity.
    }
    \label{fig:training_loss_alpha}
    \vspace{-1em}
\end{figure}

\paragraph{Training Dynamics under Different Learnability Penalties.}
We further examine the training dynamics of student models under different strengths of the learnability penalty in MI-Distillation. 
Figure~\ref{fig:training_loss_alpha} plots the training loss of Qwen2.5-3B-Instruct and Llama-3.2-3B-Instruct when varying the penalty coefficient $\alpha$.
Across both student backbones, MI-Distillation exhibits substantially lower and more stable training loss than Long CoT and Mix Long, indicating that selecting trajectories with explicit learnability control alleviates the optimization difficulty introduced by overly complex reasoning traces.
Meanwhile, increasing $\alpha$ generally leads to smoother optimization and lower loss, especially for Llama-3.2-3B-Instruct, suggesting that stronger learnability-aware selection can better align the distilled trajectories with the student model's learning capacity.
Although Short CoT achieves the lowest training loss, it provides less informative reasoning supervision. MI-Distillation maintains a favorable balance by preserving richer reasoning signals while remaining more learnable than direct Long-CoT distillation.

\section{Conclusion}
In this work, we investigate Long CoT distillation for small student models from a gradient-centric perspective. Our analysis shows that Long CoT supervision induces larger and more concentrated gradient updates than Short CoT, with the effect becoming stronger as model capacity increases. These results suggest that effective reasoning distillation requires balancing rationale informativeness with student learnability. Based on this insight, we propose \textbf{MI-Distillation}, which constructs a continuous Instruct-Reasoning data spectrum via model interpolation, and \textbf{SeqLSS}, a student-aware criterion for selecting informative and learnable trajectories. Extensive experiments demonstrate that MI-Distillation consistently improves small-model CoT distillation over strong baselines, highlighting a practical path toward transferring reasoning abilities from LRMs to compact models.

\section*{Limitations}
While MI-Distillation with SeqLSS shows consistent benefits, several limitations remain. 
First, our experiments mainly focus on mathematical reasoning benchmarks, where final-answer verification is relatively reliable. Extending MI-Distillation to open-ended reasoning, code generation, multilingual tasks, and domains with weaker automatic verification remains an important direction for future work. Second, SeqLSS requires scoring multiple candidate rationales with the student model, introducing additional offline data-construction cost compared with distillation from a single teacher. While this cost is incurred before training, more efficient scoring or approximation strategies would further improve scalability to larger corpora and broader teacher spectra. Finally, SeqLSS estimates student-side learnability through token-level likelihoods, which may not fully capture higher-level reasoning properties such as logical faithfulness, conciseness, robustness, or step-level correctness. Future work could integrate learnability-aware selection with process-level verification, step-wise feedback, or stronger rationale-quality assessment.

\section*{Acknowledgments}
This research is supported by the National Natural Science Foundation of China (No.62476127,62306140), the Natural Science Foundation of Jiangsu Province (No.BK20242039), the Fundamental Research Fund for the Central Universities (NO.NXD2026006), the Disciplinary Basic Research Project Fund (No.ILF26023), the Research Fund (No.PO250624101698), the Scientific Research Starting Foundation of Nanjing University of Aeronautics and Astronautics (No.YQR21022), and the High Performance Computing Platform of Nanjing University of Aeronautics and Astronautics.

\bibliography{custom}

\appendix

\section{Detailed Problem Setup}
\label{sec:appendix_problem_setup}

We study chain-of-thought reasoning distillation from a large teacher model to a compact student model.
Given a supervised reasoning dataset
$\mathcal{D}=\{(q_i,a_i)\}_{i=1}^{N}$, where $q_i$ denotes the input problem and $a_i$ denotes the reference answer, a teacher model $\theta_T$ is prompted to generate a reasoning trajectory $R_i$ conditioned on $q_i$.
Each trajectory contains intermediate reasoning steps together with a final prediction, and serves as process-level supervision for student fine-tuning.

The resulting distillation corpus is constructed as
\[
\mathcal{D}_{\mathrm{distill}}
=
\{(q_i, R_i)\}_{i=1}^{N}.
\]
The student model $\theta_S$ is then optimized with the standard next-token prediction objective:
\[
\mathcal{L}(\theta_S)
=
-\sum_{(q_i,R_i)\in \mathcal{D}_{\mathrm{distill}}}
\log p_{\theta_S}(R_i \mid q_i).
\]

Process-level supervision goes beyond final-answer imitation by exposing the student to explicit reasoning trajectories.
However, teacher reasoning trajectories can differ substantially in their utility for compact students: more detailed rationales may encode richer reasoning information, but may also introduce greater learning difficulty.
The tension between reasoning informativeness and student learnability motivates our study of how reasoning supervision interacts with student capacity, as well as the student-aware trajectory selection strategy in MI-Distillation.

\section{Detailed Preliminaries and Gradient Formulations}
\label{sec:appendix_gradient}

In Section \ref{sec:gradient_dynamics}, we omitted the module-specific and layer-wise notations to provide a high-level, intuitive overview of the gradient signatures. This appendix presents the complete mathematical formulations, including the optimization objective and the rigorous layer-wise definitions of the SVD-based metrics.

\subsection{Supervised Fine-Tuning Objective}
Formally, let $(x, y)$ denote a training instance from the distillation dataset, where $x$ represents the input instruction or problem, and $y = (y_1, \ldots, y_T)$ denotes the target sequence consisting of the CoT rationale followed by the final answer. Let $\mathcal{M}_\theta$ be the student model parameterized by $\theta$. For a given instance $(x, y)$, Supervised Fine-Tuning (SFT) optimizes the model by minimizing the normalized token-level negative log-likelihood:
\begin{equation}
    \mathcal{L}_{\mathrm{SFT}}(\theta; x, y) = -\frac{1}{T} \sum_{t=1}^{T} \log p_\theta(y_t \mid x, y_{<t})
\end{equation}

\subsection{Layer-wise Gradient Extraction}
To analyze the optimization behavior, we focus on the gradients associated with the attention projection modules. For the $i$-th layer of an $N$-layer student LLM, where $i \in \{0, \ldots, N-1\}$, we denote the gradients of the query, key, value, and output projection matrices as $\mathbf{G}_{Q}^{(i)}$, $\mathbf{G}_{K}^{(i)}$, $\mathbf{G}_{V}^{(i)}$, and $\mathbf{G}_{O}^{(i)}$, respectively. 

For analytical brevity, let $\mathbf{G}_{X,i} \in \mathbb{R}^{m \times n}$ represent the gradient matrix for any given projection $X \in \{Q, K, V, O\}$ at layer $i$.

\subsection{SVD-based Gradient Metrics}
Directly quantifying high-dimensional gradient matrices in modern LLMs is computationally challenging and often uninformative. Following~\cite{li2024happened}, we characterize the spectral structure of each gradient matrix $\mathbf{G}_{X,i}$ through Singular Value Decomposition (SVD)~\cite{carlini2024stealing}:
\begin{equation}
    \mathbf{G}_{X,i} = \mathbf{U}_{X,i} \mathbf{\Sigma}_{X,i} \mathbf{V}_{X,i}^{\top}
\end{equation}
where $\mathbf{\Sigma}_{X,i}$ is the diagonal matrix containing the singular values. Let $r = \min(m, n)$ denote the maximum possible rank. We summarize the resulting singular-value spectrum using two metrics:

\paragraph{Nuclear Norm.}
The nuclear norm measures the total spectral mass of the gradient, reflecting the overall magnitude of the optimization signal induced by distillation. It is formally defined as:
\begin{equation}
    \|\mathbf{G}_{X,i}\|_* = \sum_{k=1}^{r} \sigma_k^{(X,i)}
\end{equation}
where $\sigma_k^{(X,i)}$ denotes the $k$-th singular value of $\mathbf{G}_{X,i}$.

\paragraph{Effective Rank.}
The effective rank measures the Shannon entropy of the normalized singular-value spectrum. It mathematically captures the number of effective dimensions involved in the gradient update:
\begin{align}
    \operatorname{ERank}(\mathbf{G}_{X,i}) &= \exp\!\left( - \sum_{k=1}^{r} p_k^{(X,i)} \log p_k^{(X,i)} \right), \\
    \text{where} \quad p_k^{(X,i)} &= \frac{\sigma_k^{(X,i)}}{\|\mathbf{G}_{X,i}\|_*} \nonumber
\end{align}
Overall, the nuclear norm captures the magnitude of gradient updates, while the effective rank captures their spectral concentration. 
Together, they allow us to compare how Short CoT and Long CoT supervision differ not only in update strength, but also in the dimensional structure of the resulting optimization trajectories.

\section{Experimental Setup Details}
\label{sec:appendix_exp_details}

\paragraph{Models.}
We instantiate the Instruct--Reasoning spectrum at two teacher scales. 
For the 32B setting, we use QwQ-32B~\cite{qwq32b} as the reasoning-oriented teacher and Qwen2.5-32B-Instruct~\cite{qwen2.5} as the instruction-oriented teacher. 
For the 14B setting, we use DeepSeek-R1-Distill-Qwen-14B~\cite{guo2025deepseek} and Qwen2.5-14B-Instruct~\cite{qwen2.5} as the corresponding reasoning- and instruction-oriented endpoints. 
Student distillation experiments are conducted on Qwen2.5-3B-Instruct~\cite{qwen2.5} and Llama-3.2-3B-Instruct~\cite{grattafiori2024llama}.

\paragraph{Training Data Generation.}
To construct the distillation corpus, we sample 7,500 problems from the MATH training split~\cite{hendrycks2021measuring}. 
Following \citet{guo2025deepseek}, we synthesize candidate reasoning trajectories using a sampling temperature of $T=0.6$ and a nucleus sampling parameter of $p=0.95$. 
All responses are generated using the model-specific chat templates detailed in Table~\ref{tab:prompt_templates}. 
The trajectories produced by both the original and interpolated teacher models form our initial candidate pool $\mathcal{D}$. 
This pool is subsequently refined through a correctness filter and our SeqLSS-based selection mechanism.

\begin{table*}[t]
\centering
\small
\renewcommand{\arraystretch}{1.5} % 增加行距，让多行文本有呼吸感
\begin{tabularx}{\textwidth}{@{} l >{\raggedright\arraybackslash}X @{}}
\toprule
\textbf{Model Family} & \textbf{Chat Template} \\
\midrule
Qwen2.5-Instruct &
\ttfamily
\textcolor{teal!80!black}{<|im\_start|>}user\newline
\{question\}\newline
Please reason step by step, and put your final answer within \textbackslash boxed\{\}.\textcolor{teal!80!black}{<|im\_end|>}\newline
\textcolor{teal!80!black}{<|im\_start|>}assistant
\\
\midrule
Llama-3.2-Instruct &
\ttfamily
\textcolor{teal!80!black}{<|begin\_of\_text|><|start\_header\_id|>}user\textcolor{teal!80!black}{<|end\_header\_id|>}\newline
\newline
\{question\}\newline
Please reason step by step, and put your final answer within \textbackslash boxed\{\}.\textcolor{teal!80!black}{<|eot\_id|><|start\_header\_id|>}assistant\textcolor{teal!80!black}{<|end\_header\_id|>}
\\
\bottomrule
\end{tabularx}
\caption{Model-specific chat templates used for teacher trajectory generation and student-model distillation. For brevity, only the user turn and the generation prompt are shown, and the system turn inserted by the default chat template is omitted. Special control tokens are highlighted in \textcolor{teal!80!black}{teal}, and \texttt{\{question\}} denotes the placeholder for the input query.}
\label{tab:prompt_templates}
\vspace{-8pt}
\end{table*}

\paragraph{Implementation Details.}
Following the dataset construction via MI-Distillation, we fine-tune the Qwen2.5-3B-Instruct and Llama-3.2-3B-Instruct models. To maintain distribution consistency, we apply the identical model-specific prompt templates utilized during the data generation phase. Our training pipeline is built upon the LLaMA-Factory framework~\citep{zheng2024llamafactory}, integrated with DeepSpeed ZeRO-3 Offloading to optimize memory efficiency without compromising throughput. Both student models are optimized using AdamW with a learning rate of $1 \times 10^{-5}$. The training spans 3 epochs with a maximum context length of 8,192 tokens, and we maintain an effective global batch size of 32 through gradient accumulation. All experiments are conducted in \texttt{bfloat16} precision for numerical stability.

\subsection{Hardware Infrastructure}
Our experimental infrastructure comprises two distinct server nodes: one equipped with 8 NVIDIA L20 GPUs and another featuring 8 NVIDIA RTX A5000 GPUs. All models and training pipelines are implemented using the PyTorch framework. 

\subsection{Inference and Evaluation}
To ensure efficient evaluation, we deploy our fine-tuned models using the \textbf{vLLM} engine \citep{kwon2023efficient}. Following the exact hyperparameter configurations utilized during the teacher model's Chain-of-Thought (CoT) generation phase, we adopt a sampling strategy with a temperature of $T=0.6$ and nucleus sampling of top-$p=0.95$. The maximum sequence length is set to 8,192 tokens.

Regarding evaluation metrics, all results are reported using $\text{pass}@1$ accuracy. 
To obtain reliable estimates under different test-set sizes, we vary the number of independent evaluation runs. 
For benchmarks with a limited number of test instances, including AIME 2024 and AMC 2023, we repeat evaluation 16 times with independently sampled outputs. 
For larger-scale benchmarks, including GSM8K, MATH-500, GPQA-Diamond, Minerva, and OlympiadBench, we repeat evaluation 4 times. 
We report the mean accuracy across runs together with the corresponding standard deviation.

\subsection{Evaluation Datasets}
\label{evaluation_datasets}
To comprehensively assess the mathematical reasoning capabilities of our models, we conduct evaluations across a diverse set of standard benchmarks. As previously mentioned, all evaluations on these datasets are performed using the identical hyperparameters employed during the teacher model's generation process. The detailed descriptions of the selected datasets are as follows:

\begin{itemize}[leftmargin=*, nosep]
    \item \textbf{GSM8K}~\citep{cobbe2021gsm8k}: 
    A collection of high-quality grade-school math word problems that require multi-step arithmetic and linguistic reasoning.

    \item \textbf{MATH-500}~\citep{lightman2024let}: 
    A representative subset of 500 problems sampled from the MATH benchmark~\citep{hendrycks2021measuring}, spanning diverse topics and difficulty levels in competition mathematics.

    \item \textbf{AMC 2023}~\citep{AMC}: 
    Problems from the 2023 American Mathematics Competitions, designed to evaluate advanced high-school mathematical problem-solving ability.

    \item \textbf{AIME 2024}~\citep{aime24}: 
A challenging benchmark drawn from the 2024 American Invitational Mathematics Examination, evaluating advanced mathematical reasoning and problem-solving ability.

    \item \textbf{GPQA-Diamond}~\citep{rein2023gpqa}: 
    A highly challenging subset of GPQA consisting of expert-written, graduate-level multiple-choice questions in biology, physics, and chemistry, designed to test scientific reasoning beyond easily searchable knowledge.

    \item \textbf{Minerva}~\citep{lewkowycz2022solving}: 
    A suite of challenging quantitative reasoning problems covering mathematics and STEM topics, intended to assess models' ability to solve complex technical questions.

    \item \textbf{OlympiadBench}~\citep{he2024olympiadbench}: 
    A rigorous benchmark of Olympiad-level mathematics problems that evaluates high-difficulty reasoning, symbolic manipulation, and logical deduction.
\end{itemize}

\paragraph{Baseline Implementations.}
We provide further details regarding the construction of the baselines evaluated in our experiments. 
To ensure a fair and rigorous comparison, all baseline models are trained on an identical set of questions, with strict control over answer filtering, prompt formatting, student model architectures, and optimization hyperparameters, fully aligning with our proposed MI-Distillation. 
Consequently, these baselines differ solely in their underlying strategies for selecting, mixing, and scheduling the distillation trajectories.

\paragraph{Baseline Details.}
We provide implementation details for the stronger baselines considered in our experiments. 
All baselines are constructed on the same question set and trained with the same formatting, answer filtering, student models, and optimization hyperparameters as MI-Distillation.

\textbf{Mix Long}~\citep{li2025small} randomly mixes Short- and Long-CoT trajectories with a fixed $4{:}1$ ratio, serving as a simple data-level combination of concise and elaborated reasoning supervision.

\textbf{Mix Large}~\citep{li2025small} follows the same mixing ratio but incorporates teacher-scale diversity. 
Specifically, we use \texttt{Qwen2.5-32B-Instruct} as the large teacher and \texttt{Qwen2.5-14B-Instruct} as the smaller teacher to generate mixed instruction-style reasoning trajectories.

\textbf{Curriculum Learning}~\citep{jiang2025teach} is implemented strictly following the original two-stage curriculum. 
Stage~1 trains the student with the \textsc{Instruct}+\textsc{NoThink} mixture, and Stage~2 continues training from the Stage~1 checkpoint using the \textsc{NoThink}+\textsc{NoRethink} mixture. 
This baseline evaluates whether a manually scheduled transition among different reasoning formats can improve reasoning distillation.

\section{Extended Experimental Results and Analysis}
\label{sec:extended_results}

\subsection{Evaluation of Interpolated Teachers on MATH-500.}
Table~\ref{tab:teacher_math500_results} reports the performance of interpolated teacher models on MATH-500 across different difficulty levels, including both the Instruct and Reasoning endpoints. As the interpolation coefficient increases, the model places more weight on the Reasoning model, generally leading to higher accuracy and longer reasoning trajectories. This trend suggests that model interpolation induces a gradual transition from concise instruction-following behavior to more elaborate reasoning behavior.

Notably, the interpolated teachers do not form a perfectly monotonic spectrum across all difficulty levels. We attribute this to minor discrepancies between the Instruct and Reasoning endpoints, such as differences in prompt formatting, chat templates, and RoPE scaling configurations, which may introduce marginal effects, especially on easier problems. Overall, however, the interpolated teachers exhibit a clear progressive pattern, and this spectrum becomes more pronounced as problem difficulty increases.

\subsection{Comparison between Direct Interpolation and TIES-Merging}
\label{app:ties_comparison}

Our framework constructs the Instruct-Reasoning spectrum via direct parameter interpolation, as defined in Eq.~\eqref{eq:mi-interp}.
A natural concern is that such naive parameter combination may introduce interference between the two endpoint models, and that more sophisticated merging schemes may be necessary.
To examine whether our findings depend on the specific merging strategy, we compare direct interpolation with TIES-Merging~\cite{yadav2023ties} using QwQ-32B and Qwen2.5-32B-Instruct under matched interpolation ratios, and evaluate the resulting merged teachers on five reasoning benchmarks.
The results are reported in Table~\ref{tab:ties_comparison}.

TIES-Merging does not consistently outperform direct interpolation: it achieves better average performance at larger instruct weights ($\lambda \in \{0.6, 0.8\}$) but performs worse at smaller ones ($\lambda \in \{0.2, 0.4\}$).
More importantly, both merging strategies preserve the overall behavior of the interpolated teachers across the spectrum, indicating that the effectiveness of MI-Distillation is not specific to naive linear interpolation.
These results support direct interpolation as a simple and competitive design choice rather than a universally optimal one, and more advanced merging methods can be seamlessly incorporated into our framework when beneficial.

\subsection{Full Student Distillation Results with Fixed Interpolation Coefficients}
\label{sec:appendix_fixed_lambda_distillation}

Table~\ref{tab:fixed_lambda_distillation} provides the full student distillation results using fixed interpolation coefficients. 
This experiment complements the main results by evaluating how different points on the Instruct-Reasoning teacher spectrum affect downstream student performance. 
Here, $\lambda$ denotes the interpolation weight toward the instruct model, where $\lambda=1.0$ corresponds to the pure instruct teacher and $\lambda=0.0$ corresponds to the pure reasoning teacher. 
In contrast to MI-Distillation, which selects trajectories adaptively for each training instance, this setting distills training examples from a single fixed interpolated teacher.

Across both Qwen2.5-3B-Instruct and Llama3.2-3B-Instruct, intermediate interpolation coefficients generally outperform the two endpoints. 
For Qwen2.5-3B-Instruct, the best average performance is obtained at $\lambda=0.8$, while Llama3.2-3B-Instruct performs best at $\lambda=0.6$. 
These results indicate that moderate interpolation can provide more suitable supervision than either purely instruct-style or purely reasoning-style trajectories. 
At the same time, the optimal coefficient varies across student families and benchmarks, suggesting that a fixed global interpolation point is insufficient. 
This observation further supports our instance-level selection strategy in MI-Distillation, which adaptively balances reasoning informativeness with student-side learnability.

\subsection{Embedding-space analysis of the interpolated CoT spectrum.}
To further verify that model interpolation induces a meaningful transition between Long-CoT and Short-CoT supervision, we analyze the generated trajectories in the embedding space.
For each interpolation coefficient $\lambda$, we embed the corresponding CoT trajectories and visualize them using PCA.
We also compute pairwise cosine distances between trajectories generated from different interpolation coefficients.

\begin{figure*}[t]
    \centering
    \includegraphics[width=0.92\linewidth]{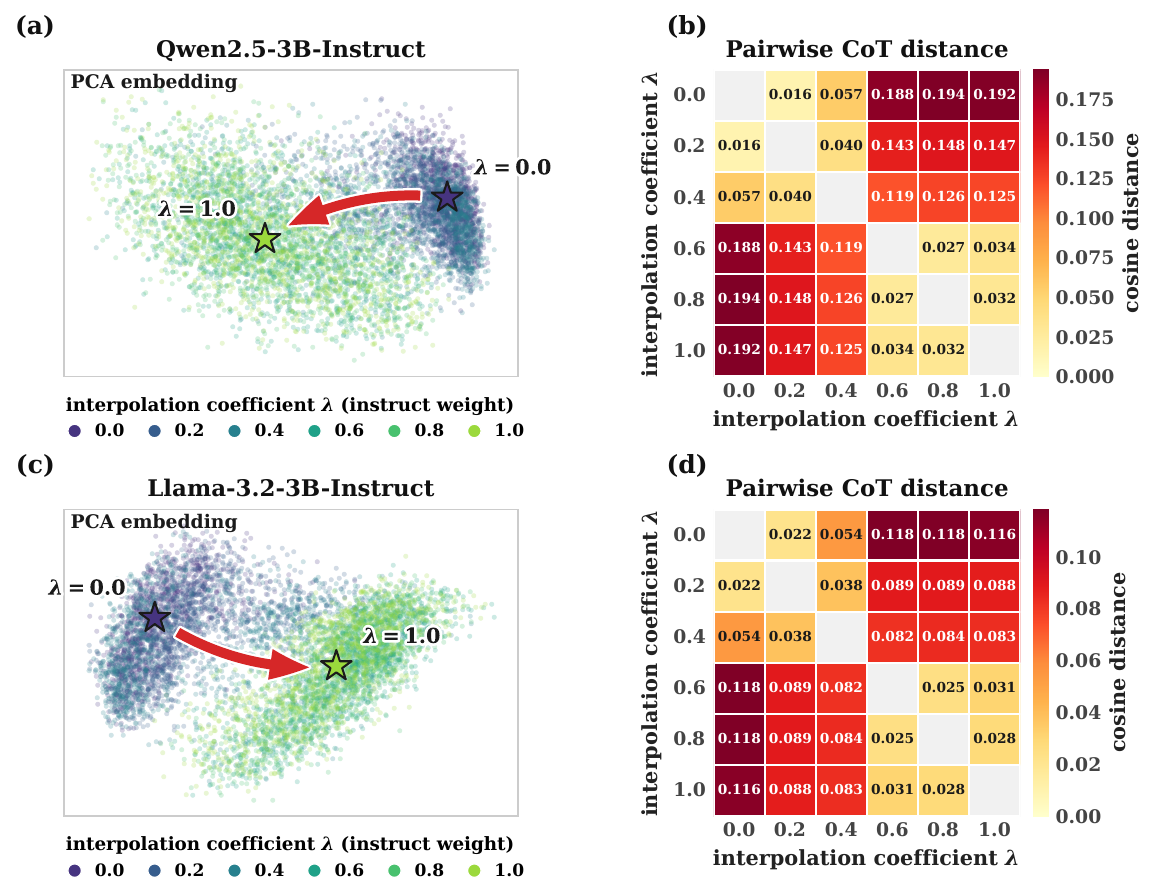}
    \caption{Visualization of the CoT representation spectrum induced by model interpolation.
    Panels (a) and (c) show PCA projections of CoT embeddings using Qwen2.5-3B-Instruct and Llama-3.2-3B-Instruct, respectively, with colors denoting the interpolation coefficient $\lambda$ (instruct weight).
    Red arrows indicate the dominant shift across the spectrum.
    Panels (b) and (d) report pairwise cosine distances between coefficients.
    Adjacent coefficients are consistently closer than distant ones, indicating that interpolation yields a gradual transition between Long-CoT and Short-CoT supervision rather than disconnected teacher distributions.}
    \label{fig:cot_embedding_pca}
\end{figure*}

As shown in Figure~\ref{fig:cot_embedding_pca}, the CoT embeddings exhibit a clear and progressive shift as $\lambda$ varies from the reasoning-oriented endpoint to the instruction-oriented endpoint.
For both Qwen2.5-3B-Instruct and Llama-3.2-3B-Instruct, neighboring interpolation coefficients tend to occupy adjacent regions in the PCA space, while endpoints are more clearly separated.
The pairwise distance matrices further support this observation: trajectories generated from nearby coefficients have smaller cosine distances, whereas trajectories from distant coefficients, especially across the two endpoints, show substantially larger distances.
These results indicate that interpolation does not simply produce isolated teacher variants, but instead constructs a smooth Instruct-Reasoning data spectrum that bridges Short CoT and Long CoT supervision.

\subsection{Ablation on the Learnability Penalty Coefficient}
\label{app:alpha_ablation}

We ablate the learnability penalty coefficient $\alpha$ in SeqLSS, which balances reasoning informativeness against student-side learnability during trajectory selection. 
As shown in Table~\ref{tab:alpha_ablation}, $\alpha$ does not induce a monotonic trend; instead, different values correspond to different trade-offs over the interpolation spectrum. 
Across both Qwen2.5-3B-Instruct and Llama-3.2-3B-Instruct, $\alpha=4$ achieves the best average performance, indicating a favorable balance between selecting informative CoT trajectories and maintaining sufficient alignment with the student model.

Smaller values of $\alpha$ remain competitive on several individual benchmarks, suggesting that the preferred strength of learnability regularization can be task- and model-dependent. 
However, $\alpha=4$ provides the most reliable overall behavior across the two student families. 
We therefore use $\alpha=4$ as the default setting in our main experiments.

\begin{table*}[t]
\centering
\small
\setlength{\tabcolsep}{5pt}
\renewcommand{\arraystretch}{1.12}
\resizebox{\textwidth}{!}{
\begin{tabular}{c l ccccc c c}
\toprule
\multirow{2}{*}{\textbf{Interp. Coef.} $\bm{\lambda}$}
& \multirow{2}{*}{\textbf{Merging}}
& \multicolumn{5}{c}{\textbf{Evaluation Datasets}}
& \multirow{2}{*}{\textbf{Avg.}}
& \multirow{2}{*}{\textbf{$\Delta$ Avg.}} \\
\cmidrule(lr){3-7}
& & \textbf{GPQA-Diamond} & \textbf{GSM8K} & \textbf{MATH-500} & \textbf{Minerva} & \textbf{Olympiad} & & \\
\midrule
\multirow{2}{*}{$\lambda=0.2$}
& Direct & $\mathbf{63.64}$ & $96.13$ & $\mathbf{90.00}$ & $\mathbf{34.93}$ & $\mathbf{62.46}$ & $\mathbf{69.43}$ & \multirow{2}{*}{$-3.12$} \\
& TIES   & $57.58$ & $\mathbf{96.44}$ & $87.20$ & $32.35$ & $58.01$ & $66.32$ & \\
\midrule
\multirow{2}{*}{$\lambda=0.4$}
& Direct & $\mathbf{54.55}$ & $\mathbf{96.06}$ & $\mathbf{88.80}$ & $\mathbf{35.29}$ & $\mathbf{58.90}$ & $\mathbf{66.72}$ & \multirow{2}{*}{$-1.27$} \\
& TIES   & $53.54$ & $95.68$ & $86.80$ & $33.82$ & $57.42$ & $65.45$ & \\
\midrule
\multirow{2}{*}{$\lambda=0.6$}
& Direct & $55.05$ & $\mathbf{96.06}$ & $83.60$ & $30.88$ & $53.12$ & $63.74$ & \multirow{2}{*}{$+2.12$} \\
& TIES   & $\mathbf{58.59}$ & $95.98$ & $\mathbf{87.80}$ & $\mathbf{31.62}$ & $\mathbf{55.34}$ & $\mathbf{65.87}$ & \\
\midrule
\multirow{2}{*}{$\lambda=0.8$}
& Direct & $\mathbf{55.05}$ & $96.06$ & $83.40$ & $30.88$ & $49.55$ & $62.99$ & \multirow{2}{*}{$+1.17$} \\
& TIES   & $53.54$ & $\mathbf{96.21}$ & $\mathbf{84.60}$ & $\mathbf{31.25}$ & $\mathbf{55.19}$ & $\mathbf{64.16}$ & \\
\bottomrule
\end{tabular}
}
\caption{
\textbf{Comparison between direct parameter interpolation and TIES-Merging for constructing interpolated teachers.}
Teachers are merged from QwQ-32B and Qwen2.5-32B-Instruct, where $\lambda$ denotes the interpolation weight toward the instruct model as in Eq.~\eqref{eq:mi-interp}.
$\Delta$~Avg.\ denotes the difference in average performance (TIES $-$ Direct).
Within each $\lambda$ group, the better result of the two merging strategies on each benchmark is shown in \textbf{bold}.
}
\label{tab:ties_comparison}
\end{table*}

\begin{table*}[!t]
\centering
\small
\setlength{\tabcolsep}{5pt} % 微调列间距以适应新增的两列，防止溢出页面边缘
\renewcommand{\arraystretch}{1.15}
\begin{tabular}{c *{12}{c}}
\toprule
\multirow{2}{*}{\textbf{MI}}
& \multicolumn{2}{c}{\textbf{Level 1}}
& \multicolumn{2}{c}{\textbf{Level 2}}
& \multicolumn{2}{c}{\textbf{Level 3}}
& \multicolumn{2}{c}{\textbf{Level 4}}
& \multicolumn{2}{c}{\textbf{Level 5}} 
& \multicolumn{2}{c}{\textbf{Average}} \\
\cmidrule(lr){2-3}\cmidrule(lr){4-5}\cmidrule(lr){6-7}\cmidrule(lr){8-9}\cmidrule(lr){10-11}\cmidrule(lr){12-13}
& \textbf{Acc.}$\uparrow$ & \textbf{Tok.}$\downarrow$
& \textbf{Acc.}$\uparrow$ & \textbf{Tok.}$\downarrow$
& \textbf{Acc.}$\uparrow$ & \textbf{Tok.}$\downarrow$
& \textbf{Acc.}$\uparrow$ & \textbf{Tok.}$\downarrow$
& \textbf{Acc.}$\uparrow$ & \textbf{Tok.}$\downarrow$ 
& \textbf{Acc.}$\uparrow$ & \textbf{Tok.}$\downarrow$ \\
\midrule
\rowcolor{rowgray}
\multicolumn{13}{c}{\textit{\textbf{14B Models}}} \\
\midrule
0.0 & 95.35 & 1,351 & \ha{94.44} & 1,465 & 94.29 & 2,097 & \ha{92.19} & 3,440 & \ha{81.34} & 4,731 & \ha{91.52} & 2,617 \\
0.2 & 95.35 & 973   & 93.33 & 1,117 & \ha{96.19} & 1,705 & 89.06 & 2,243 & 77.61 & 3,664 & 90.31 & 1,940 \\
0.4 & 93.02 & 529   & 86.67 & 644   & 91.43 & 1,101 & 79.69 & 1,521 & 71.64 & 2,609 & 84.49 & 1,281 \\
0.6 & \ha{97.67} & 318   & 93.33 & 424   & 95.24 & 588   & 78.91 & 835   & 70.15 & 1,444 & 87.06 & 722   \\
0.8 & 95.35 & \hc{293}   & 91.11 & \hc{378}   & 91.43 & 581   & 71.88 & 717   & 55.97 & 851   & 81.15 & 564   \\
1.0 & 95.35 & 348   & 93.33 & 432   & 89.52 & \hc{499}   & 75.00 & \hc{621}   & 54.48 & \hc{759}   & 81.54 & \hc{532}   \\
\midrule
\rowcolor{rowgray}
\multicolumn{13}{c}{\textit{\textbf{32B Models}}} \\
\midrule
0.0 & \ha{95.35} & 1,661 & 96.67 & 2,137 & 95.24 & 2,864 & 88.28 & 3,775 & 75.37 & 5,228 & 90.18 & 3,133 \\
0.2 & 95.35 & 1,195 & \ha{97.78} & 1,400 & 96.19 & 1,813 & \ha{91.41} & 2,533 & \ha{83.58} & 3,763 & \ha{92.86} & 2,141 \\
0.4 & 95.35 & 499   & 94.44 & 752   & \ha{97.14} & 930   & 87.50 & 1,483 & 82.84 & 2,298 & 91.45 & 1,192 \\
0.6 & 95.35 & 362   & 94.44 & 489   & 95.24 & 543   & 81.25 & 742   & 73.13 & 922   & 87.88 & 612   \\
0.8 & 93.02 & 353   & 91.11 & 441   & 93.33 & 515   & 81.25 & 669   & 67.91 & 897   & 85.32 & 575   \\
1.0 & 95.35 & \hc{339}   & 91.11 & \hc{425}   & 92.38 & \hc{502}   & 78.91 & \hc{600}   & 64.93 & \hc{714}   & 84.54 & \hc{516}   \\
\bottomrule
\end{tabular}
\caption{Performance of the interpolated teacher model on \textsc{MATH-500} across difficulty levels (Level~1--5).
\textbf{$\lambda$} denotes the model-interpolation coefficient, ranging from the reasoning model ($\lambda=0$) to the instruct model ($\lambda=1$).
For each level and the \textbf{average} performance, we report accuracy (\textbf{Acc.}) and generated tokens (\textbf{Tok.}).
Within each model-size group, the \textbf{best accuracy} and the \textbf{lowest token cost} per column are shown in \colorbox{hlblue}{\textbf{bold}} and \colorbox{hltok}{\textbf{bold}}.}
\label{tab:teacher_math500_results}
\end{table*}

\begin{table*}[t]
\centering
\small
\setlength{\tabcolsep}{4.0pt}
\renewcommand{\arraystretch}{1.12}
\definecolor{hlblue}{RGB}{230, 241, 252}

\resizebox{\textwidth}{!}{
\begin{tabular}{l cccccccc}
\toprule
\multirow{2}{*}{\textbf{Interp. Coef.} $\boldsymbol{\lambda}$}
& \multicolumn{7}{c}{\textbf{Evaluation Datasets}}
& \multirow{2}{*}{\textbf{Avg.}} \\
\cmidrule(lr){2-8}
& \textbf{AIME24}
& \textbf{AMC23}
& \textbf{GPQA-Diamond}
& \textbf{GSM8K}
& \textbf{MATH-500}
& \textbf{Minerva}
& \textbf{Olympiad}
& \\

\midrule
\rowcolor{gray!10}
\multicolumn{9}{c}{\textit{\textbf{Student Model: Qwen2.5-3B-Instruct}}} \\
\midrule

$\lambda=1.0$
& $5.42_{\pm 1.67}$
& $35.00_{\pm 6.65}$
& $25.88_{\pm 2.87}$
& $83.49_{\pm 1.04}$
& $62.20_{\pm 1.97}$
& $25.37_{\pm 1.59}$
& $27.00_{\pm 0.94}$
& $37.77$ \\

$\lambda=0.8$
& $5.62_{\pm 2.91}$
& $37.19_{\pm 4.82}$
& \cellcolor{hlblue}$\mathbf{28.03}_{\pm 0.65}$
& \cellcolor{hlblue}$\mathbf{84.63}_{\pm 0.83}$
& $63.20_{\pm 1.26}$
& $26.75_{\pm 1.91}$
& $29.23_{\pm 1.62}$
& \cellcolor{hlblue}$\mathbf{39.24}$ \\

$\lambda=0.6$
& $3.33_{\pm 2.72}$
& \cellcolor{hlblue}$\mathbf{39.22}_{\pm 6.10}$
& $25.88_{\pm 3.48}$
& $84.48_{\pm 0.23}$
& $63.20_{\pm 1.10}$
& \cellcolor{hlblue}$\mathbf{27.21}_{\pm 1.44}$
& \cellcolor{hlblue}$\mathbf{30.19}_{\pm 1.84}$
& $39.07$ \\

$\lambda=0.4$
& \cellcolor{hlblue}$\mathbf{6.04}_{\pm 3.04}$
& $36.88_{\pm 4.23}$
& $25.63_{\pm 2.04}$
& $84.00_{\pm 0.77}$
& \cellcolor{hlblue}$\mathbf{63.45}_{\pm 0.62}$
& $24.72_{\pm 1.21}$
& $28.56_{\pm 1.46}$
& $38.47$ \\

$\lambda=0.2$
& $3.75_{\pm 2.69}$
& $32.34_{\pm 4.96}$
& $23.86_{\pm 2.68}$
& $83.07_{\pm 1.13}$
& $58.70_{\pm 1.05}$
& $22.24_{\pm 1.26}$
& $24.89_{\pm 1.05}$
& $35.55$ \\

$\lambda=0.0$
& $3.75_{\pm 3.19}$
& $27.03_{\pm 4.10}$
& $19.32_{\pm 0.96}$
& $80.78_{\pm 0.43}$
& $54.05_{\pm 0.72}$
& $20.13_{\pm 1.25}$
& $21.11_{\pm 1.18}$
& $32.31$ \\

\midrule
\rowcolor{gray!10}
\multicolumn{9}{c}{\textit{\textbf{Student Model: Llama3.2-3B-Instruct}}} \\
\midrule

$\lambda=1.0$
& $3.96_{\pm 2.78}$
& $24.38_{\pm 5.88}$
& $23.99_{\pm 2.35}$
& $76.65_{\pm 1.17}$
& $51.40_{\pm 1.25}$
& $17.10_{\pm 2.43}$
& $17.43_{\pm 1.44}$
& $30.70$ \\

$\lambda=0.8$
& \cellcolor{hlblue}$\mathbf{6.25}_{\pm 3.42}$
& \cellcolor{hlblue}$\mathbf{30.16}_{\pm 5.51}$
& $24.75_{\pm 3.27}$
& $78.37_{\pm 0.53}$
& $50.80_{\pm 1.66}$
& $17.00_{\pm 1.06}$
& $19.96_{\pm 1.24}$
& $32.47$ \\

$\lambda=0.6$
& \cellcolor{hlblue}$\mathbf{6.25}_{\pm 3.42}$
& $28.59_{\pm 4.65}$
& $25.13_{\pm 2.52}$
& \cellcolor{hlblue}$\mathbf{79.34}_{\pm 0.59}$
& \cellcolor{hlblue}$\mathbf{52.45}_{\pm 2.18}$
& $16.82_{\pm 0.76}$
& \cellcolor{hlblue}$\mathbf{20.47}_{\pm 0.63}$
& \cellcolor{hlblue}$\mathbf{32.72}$ \\

$\lambda=0.4$
& $5.21_{\pm 3.65}$
& $27.19_{\pm 5.69}$
& \cellcolor{hlblue}$\mathbf{26.39}_{\pm 1.51}$
& $78.92_{\pm 1.21}$
& $51.35_{\pm 1.81}$
& \cellcolor{hlblue}$\mathbf{17.46}_{\pm 1.42}$
& $18.29_{\pm 1.24}$
& $32.12$ \\

$\lambda=0.2$
& $3.54_{\pm 2.57}$
& $23.91_{\pm 2.41}$
& $23.11_{\pm 2.81}$
& $78.34_{\pm 0.76}$
& $49.60_{\pm 0.85}$
& $15.44_{\pm 1.27}$
& $15.21_{\pm 0.89}$
& $29.88$ \\

$\lambda=0.0$
& $0.83_{\pm 1.49}$
& $20.00_{\pm 4.18}$
& $22.47_{\pm 2.69}$
& $78.53_{\pm 0.97}$
& $45.25_{\pm 2.46}$
& $13.24_{\pm 1.08}$
& $13.87_{\pm 1.00}$
& $27.74$ \\

\bottomrule
\end{tabular}
}

\caption{
Full student distillation results using fixed interpolation coefficients.
Here, $\lambda$ denotes the interpolation weight toward the instruct model, with $\lambda=1.0$ corresponding to the pure instruct teacher and $\lambda=0.0$ to the pure reasoning teacher.
All results are reported as Pass@1 accuracy, averaged over multiple evaluation runs with standard deviations.
Specifically, AIME24 and AMC23 are evaluated over 16 runs due to their limited test-set sizes, while the remaining benchmarks are evaluated over 4 runs.
The best result within each student-model block is highlighted.
}
\label{tab:fixed_lambda_distillation}
\end{table*}

\begin{table*}[t]
\centering
\small
\setlength{\tabcolsep}{4.2pt}
\renewcommand{\arraystretch}{1.12}
\definecolor{hlblue}{RGB}{230, 241, 252}

\resizebox{\textwidth}{!}{
\begin{tabular}{c ccc ccc cc}
\toprule
\multirow{2}{*}{\textbf{Coefficient}} 
& \multicolumn{7}{c}{\textbf{Evaluation Datasets}} 
& \multirow{2}{*}{\textbf{Avg.}} \\
\cmidrule(lr){2-8}
& \textbf{AIME24} 
& \textbf{AMC23} 
& \textbf{GPQA-Diamond} 
& \textbf{GSM8K} 
& \textbf{MATH-500} 
& \textbf{Minerva} 
& \textbf{Olympiad} 
& \\

\midrule
\rowcolor{gray!10}
\multicolumn{9}{c}{\textit{\textbf{Student Model: Qwen2.5-3B-Instruct}}} \\
\midrule

$\alpha=1$
& $\underline{5.62}_{\pm 2.91}$
& $34.53_{\pm 5.93}$
& $\mathbf{30.43}_{\pm 5.29}$
& $\mathbf{85.33}_{\pm 0.27}$
& $\underline{64.92}_{\pm 1.74}$
& $25.18_{\pm 2.66}$
& $\mathbf{30.08}_{\pm 1.64}$
& 39.44 \\

$\alpha=2$
& $\mathbf{6.04}_{\pm 2.50}$
& $\underline{39.84}_{\pm 4.70}$
& $29.42_{\pm 1.45}$
& $84.61_{\pm 0.47}$
& $64.60_{\pm 0.86}$
& $\mathbf{27.02}_{\pm 1.36}$
& $29.78_{\pm 1.18}$
& $\underline{40.19}$ \\

$\alpha=3$
& $\mathbf{6.04}_{\pm 3.89}$
& $37.81_{\pm 4.27}$
& $\underline{30.05}_{\pm 0.65}$
& $\underline{85.24}_{\pm 0.45}$
& $64.75_{\pm 1.62}$
& $24.26_{\pm 0.74}$
& $\underline{29.93}_{\pm 1.18}$
& 39.73 \\

\rowcolor{hlblue}
\textbf{$\alpha=4$}
& $5.42_{\pm 2.06}$
& $\mathbf{40.31}_{\pm 5.39}$
& $29.55_{\pm 1.27}$
& $85.06_{\pm 0.53}$
& $\mathbf{66.15}_{\pm 1.08}$
& $\underline{25.46}_{\pm 1.10}$
& $29.49_{\pm 0.81}$
& $\mathbf{40.21}$ \\

\midrule
\rowcolor{gray!10}
\multicolumn{9}{c}{\textit{\textbf{Student Model: Llama-3.2-3B-Instruct}}} \\
\midrule

$\alpha=1$
& $4.79_{\pm 2.71}$
& $25.16_{\pm 3.82}$
& $23.86_{\pm 3.31}$
& $\underline{77.65}_{\pm 0.50}$
& $52.25_{\pm 0.82}$
& $16.45_{\pm 3.20}$
& $19.14_{\pm 1.21}$
& 31.33 \\

$\alpha=2$
& $5.42_{\pm 2.69}$
& $27.50_{\pm 5.08}$
& $21.97_{\pm 4.67}$
& $77.60_{\pm 0.60}$
& $\underline{52.85}_{\pm 1.00}$
& $\underline{17.10}_{\pm 1.14}$
& $19.55_{\pm 0.88}$
& 31.71 \\

$\alpha=3$
& $\underline{6.04}_{\pm 3.49}$
& $\mathbf{30.63}_{\pm 4.87}$
& $\mathbf{26.39}_{\pm 2.08}$
& $77.60_{\pm 0.80}$
& $51.70_{\pm 1.41}$
& $\mathbf{18.29}_{\pm 1.35}$
& $\underline{19.58}_{\pm 0.51}$
& $\underline{32.89}$ \\

\rowcolor{hlblue}
\textbf{$\alpha=4$}
& $\mathbf{6.88}_{\pm 3.10}$
& $\underline{30.47}_{\pm 2.28}$
& $\underline{24.62}_{\pm 2.52}$
& $\mathbf{77.75}_{\pm 0.77}$
& $\mathbf{54.60}_{\pm 1.02}$
& $16.27_{\pm 0.46}$
& $\mathbf{19.96}_{\pm 0.28}$
& $\mathbf{32.93}$ \\

\bottomrule
\end{tabular}
}
\caption{\textbf{Ablation study on the learnability penalty coefficient $\alpha$ in MI-Distillation.}
We evaluate SeqLSS-based trajectory selection under different values of $\alpha$ on two 3B student models.
For each student-model block, the best result in each column is shown in \textbf{bold}, and the second-best result is \underline{underlined}.
The default setting, $\alpha=4$, is highlighted in blue.
Standard deviations are reported as subscripts for compactness.}
\label{tab:alpha_ablation}
\vspace{-1em}
\end{table*}

\clearpage
\onecolumn

\section{Examples of Model Outputs with Different Interpolation Weights}
\label{sec:appendix_interpolation_examples}

\begin{promptbox}
Joe is studying a bacteria population. There are 20 bacteria present at 3:00 p.m. and the population doubles every 3 minutes. Assuming none of the bacteria die, how many bacteria are present at 3:15 p.m. the same day?

Please reason step by step, and put your final answer within \verb|\boxed{}|.
\end{promptbox}

\begin{modelbox}{$\lambda = 0.0$ (Reasoning Model)}
Okay, let's see. Joe is studying bacteria that double every 3 minutes. Starting at 3:00 p.m. with 20 bacteria, and we need to find out how many there are at 3:15 p.m. Hmm, so first, how much time passes between 3:00 and 3:15? That's 15 minutes. Right. Since the bacteria double every 3 minutes, I need to figure out how many 3-minute intervals are in 15 minutes. Let me do that division: 15 divided by 3 equals 5. So there are 5 doubling periods.

Now, starting with 20 bacteria, each doubling period multiplies the population by 2. So after each 3 minutes, the number doubles. So after the first 3 minutes, it's 20 * 2 = 40. Then after 6 minutes, that's 40 * 2 = 80. Continuing this way: 3rd interval (9 minutes total) would be 160, 4th interval (12 minutes) 320, and the 5th interval (15 minutes) would be 640. Wait, but maybe there's a formula for exponential growth here?

The formula is $N = N0 * 2^{(t/T)}$, where N0 is initial amount, t is total time, and T is the doubling period. Let me plug in the numbers. N0 is 20, t is 15 minutes, T is 3 minutes. So that's $20 * 2^{(15/3)} = 20 * 2^5$. $2^5$ is 32, so 20 * 32 = 640. Hmm, same as before. So that seems right. Let me check again. 15 divided by 3 is indeed 5, so $2^5$ is 32. 20 times 32 is 640. Yeah, that makes sense. So the answer should be 640. Wait, is there a chance I made a mistake in the formula? Let me think. The formula for exponential growth when doubling is indeed $N = N0 * 2^{(t/\text{doubling time})}$. So yes, that's correct. Alternatively, step by step:

At 3:00 pm: 20

3:03 pm: 40

3:06 pm: 80

3:09 pm: 160

3:12 pm: 320

3:15 pm: 640

Yep, that's five doublings. Each time multiplying by 2. So $2^5$ is 32, times 20 gives 640. I think that's right. So the final answer is 640. Let me just make sure I didn't miscalculate $2^5$. 2, 4, 8, 16, 32. Yes, five doublings. So 20*32 is indeed 640. Okay, confident now.

The bacteria population doubles every 3 minutes. Starting at 3:00 p.m. with 20 bacteria, we need to calculate the population at 3:15 p.m., which is 15 minutes later. 

First, determine the number of doubling periods in 15 minutes:
\[
\frac{15 \text{ minutes}}{3 \text{ minutes per doubling}} = 5 \text{ doublings}
\]

Using the formula for exponential growth \( N = N_0 \times 2^t \), where \( N_0 = 20 \), \( t = 5 \):
\[
N = 20 \times 2^5 = 20 \times 32 = 640
\]

Thus, the number of bacteria present at 3:15 p.m. is \(\boxed{640}\).
\end{modelbox}

\begin{modelbox}{$\lambda = 0.2$}
Okay, let's see. Joe is studying bacteria that double every 3 minutes. At 3:00 p.m., there are 20 bacteria. We need to find out how many there are at 3:15 p.m. Hmm, first, I should figure out how much time passes between 3:00 and 3:15. That's 15 minutes, right? 

Now, since the bacteria double every 3 minutes, I need to determine how many 3-minute intervals are in 15 minutes. Let me divide 15 by 3. 15 divided by 3 is 5. So, there are 5 intervals of 3 minutes each in 15 minutes. 

That means the bacteria population doubles 5 times. Starting with 20 bacteria, each doubling multiplies the population by 2. So after the first 3 minutes, it would be 20 * 2 = 40. Then after 6 minutes, 40 * 2 = 80. Continuing this way:

At 3:03 p.m. - 40
At 3:06 p.m. - 80
At 3:09 p.m. - 160
At 3:12 p.m. - 320
At 3:15 p.m. - 640

Wait, but maybe there's a formula for exponential growth that I can use instead of calculating each step. The formula is $N = N0 * 2^{(t/T)}$, where N0 is the initial population, t is the total time, and T is the doubling time. 

Let me plug in the numbers. N0 is 20, t is 15 minutes, and T is 3 minutes. So $N = 20 * 2^{(15/3)}$. 15 divided by 3 is 5, so that's $2^5$. $2^5$ is 32. Then 20 * 32 equals 640. 

Hmm, both methods give me 640. So I think that's the answer. Let me just double-check the steps again. Starting at 20, doubling 5 times: 20, 40, 80, 160, 320, 640. Yep, that's five doublings. Each time multiplying by 2, so $2^5$ is 32, times 20 is 640. 

I think that's right. The key was figuring out how many doubling periods fit into 15 minutes, which is 5. Then applying the exponential growth formula. So the final answer should be 640 bacteria at 3:15 p.m.

The bacteria population doubles every 3 minutes. From 3:00 p.m. to 3:15 p.m., there are 15 minutes, which is \( \frac{15}{3} = 5 \) doubling periods. Starting with 20 bacteria, the population after 5 doublings is calculated as:
\[
20 \times 2^5 = 20 \times 32 = 640
\]

$\boxed{640}$
\end{modelbox}

\begin{modelbox}{$\lambda = 0.4$}
Okay, let's see. Joe is studying bacteria that double every 3 minutes. At 3:00 p.m., there are 20 bacteria. We need to find out how many there are at 3:15 p.m. Hmm, first, I should figure out how much time passes between 3:00 and 3:15. That's 15 minutes, right?

Now, since the bacteria double every 3 minutes, I need to see how many 3-minute intervals are in 15 minutes. Let me divide 15 by 3. That's 5. So there are 5 intervals of 3 minutes each in 15 minutes.

Each interval, the population doubles. Starting with 20 bacteria, after the first 3 minutes, it becomes 20 * 2 = 40. Then after another 3 minutes (6 minutes total), it doubles again to 40 * 2 = 80. Continuing this way:

- At 3:03 p.m. (3 minutes): 20 * 2 = 40
- At 3:06 p.m. (6 minutes): 40 * 2 = 80
- At 3:09 p.m. (9 minutes): 80 * 2 = 160
- At 3:12 p.m. (12 minutes): 160 * 2 = 320
- At 3:15 p.m. (15 minutes): 320 * 2 = 640

Alternatively, I can use the formula for exponential growth: $N = N0 * 2^{(t/T)}$, where N0 is the initial number, t is the time elapsed, and T is the doubling time. Here, N0 = 20, t = 15 minutes, T = 3 minutes. Plugging in:

$N = 20 * 2^{(15/3)} = 20 * 2^5 = 20 * 32 = 640.$

So both methods give me 640 bacteria at 3:15 p.m. Let me just check my calculations again. $2^5$ is 32, and 20 times 32 is indeed 640. Yep, that seems right.

Therefore, the final answer is $\boxed{640}$.
\end{modelbox}

\begin{modelbox}{$\lambda = 0.6$}
To determine the number of bacteria present at 3:15 p.m., we need to calculate how many times the population doubles between 3:00 p.m. and 3:15 p.m.

1. \textbf{Determine the time interval:}
   - From 3:00 p.m. to 3:15 p.m. is a 15-minute period.

2. \textbf{Calculate the number of doubling periods:}
   - The population doubles every 3 minutes.
   - Number of doubling periods = Total time interval / Doubling time
   - Number of doubling periods = 15 minutes / 3 minutes per doubling = 5 doublings

3. \textbf{Initial population:}
   - At 3:00 p.m., there are 20 bacteria.

4. \textbf{Population growth calculation:}
   - Each doubling period multiplies the population by 2.
   - After 5 doublings, the population is multiplied by \(2^5\).

5. \textbf{Final population calculation:}
   - Final population = Initial population * \(2^5\)
   - Final population = 20 * \(2^5\)
   - \(2^5 = 32\)
   - Final population = 20 * 32 = 640

Thus, the number of bacteria present at 3:15 p.m. is \(\boxed{640}\).
\end{modelbox}

\begin{modelbox}{$\lambda = 0.8$}
To determine the number of bacteria present at 3:15 p.m., given that the population doubles every 3 minutes and starts with 20 bacteria at 3:00 p.m., we can follow these steps:

1. \textbf{Identify the time interval and the doubling period:}
   - The time interval from 3:00 p.m. to 3:15 p.m. is 15 minutes.
   - The bacteria population doubles every 3 minutes.

2. \textbf{Calculate the number of doubling periods within the 15-minute interval:}
   - Number of doubling periods = Total time interval / Doubling period
   - Number of doubling periods = 15 minutes / 3 minutes per period = 5 periods

3. \textbf{Use the formula for exponential growth:}
   - The population after \( n \) doubling periods can be calculated using the formula:
     \[
     P = P_0 \times 2^n
     \]
     where \( P \) is the final population, \( P_0 \) is the initial population, and \( n \) is the number of doubling periods.

4. \textbf{Plug in the known values:}
   - Initial population \( P_0 = 20 \)
   - Number of doubling periods \( n = 5 \)
   - Final population \( P = 20 \times 2^5 \)

5. \textbf{Calculate the final population:}
   - \( 2^5 = 32 \)
   - \( P = 20 \times 32 = 640 \)

Therefore, the number of bacteria present at 3:15 p.m. is \(\boxed{640}\).
\end{modelbox}

\begin{modelbox}{$\lambda = 1.0$ (Instruct Model)}
To solve this problem, we need to calculate how many times the bacteria population doubles from 3:00 p.m. to 3:15 p.m.

1. \textbf{Identify the time span and doubling period:}
   - The time span from 3:00 p.m. to 3:15 p.m. is 15 minutes.
   - The bacteria population doubles every 3 minutes.

2. \textbf{Determine the number of doubling periods within the 15-minute span:}
   - Number of doubling periods = Total time span / Doubling period
   - Number of doubling periods = 15 minutes / 3 minutes per doubling = 5

3. \textbf{Calculate the population after each doubling period:}
   - Initial population at 3:00 p.m. = 20
   - Population doubles every 3 minutes.

4. \textbf{Use the formula for exponential growth:}
   - The population after \( n \) doubling periods is given by \( P = P_0 \times 2^n \), where \( P_0 \) is the initial population and \( n \) is the number of doubling periods.
   - Here, \( P_0 = 20 \) and \( n = 5 \).

5. \textbf{Substitute the values into the formula:}
   - \( P = 20 \times 2^5 \)

6. \textbf{Calculate the final population:}
   - \( P = 20 \times 32 \)
   - \( P = 640 \)

Thus, the number of bacteria present at 3:15 p.m. is \(\boxed{640}\).
\end{modelbox}
\end{document}